\pdfoutput=1

\documentclass[11pt]{article}

\usepackage{acl}

\usepackage{times}
\usepackage{latexsym}

\usepackage{hyperref}
\usepackage{url}
\usepackage{booktabs}
\usepackage{xcolor}
\usepackage{graphicx}
\usepackage{subcaption}
\usepackage{soul}
\usepackage[colorinlistoftodos,prependcaption,textsize=small]{todonotes}

\usepackage[T1]{fontenc}

\usepackage[utf8]{inputenc}

\usepackage{microtype}

\title{The Fine-Tuning Paradox: Boosting Translation Quality Without Sacrificing LLM Abilities}

 \author{
    {\bf David Stap$^2$\thanks{~~Work done during an internship at Amazon.} \hspace{2mm}}
    {\bf Eva Hasler$^1$ \hspace{2mm}}
    {\bf Bill Byrne$^1$ \hspace{2mm}} \\
    {\bf Christof Monz$^2$}
    {\bf Ke Tran$^1$ \hspace{2mm}} \\
    $^1$Amazon AI, Berlin \\
    $^2$Language Technology Lab, University of Amsterdam\\
    \medskip
    \texttt{\{d.stap, c.monz\}@uva.nl},
    \texttt{\{ehasler, willbyrn, trnke\}@amazon.com}
 }

\begin{document}

\maketitle

\begin{abstract}
Fine-tuning large language models (LLMs) for machine translation has shown improvements in overall translation quality.
However, it is unclear what is the impact of fine-tuning on desirable LLM behaviors that are not present in neural machine translation models, such as steerability, inherent document-level translation abilities, and the ability to produce less literal translations.
We perform an extensive translation evaluation on the LLaMA and Falcon family of models with model size ranging from 7 billion up to 65 billion parameters.
Our results show that while fine-tuning improves the general translation quality of LLMs, several abilities degrade.
In particular, we observe a decline in the ability to perform formality steering, to produce technical translations through few-shot examples, and to perform document-level translation.
On the other hand, we observe that the model produces less literal translations after fine-tuning on parallel data.
We show that by including monolingual data as part of the fine-tuning data we can maintain the abilities while simultaneously enhancing overall translation quality.
Our findings emphasize the need for fine-tuning strategies that preserve the benefits of LLMs for machine translation.
\end{abstract}

\section{Introduction}
Recent work has highlighted a range of qualitative advantages that large language models (LLMs) hold over Neural Machine Translation (NMT) models.
One significant advantage is the controllability of style and language variety which can be achieved through prompting and in-context learning \citep{brown_language_2020,garcia_unreasonable_2023,agrawal_-context_2023}.
LLMs also exhibit inherent document-level translation abilities \citep{wang_document-level_2023, karpinska_large_2023}.
Another advantage is their ability to produce less literal translations \citep{raunak_gpts_2023}.
Finally, LLMs have been shown to have better performance in handling difficult linguistic phenomena such as idioms and ambiguous expressions \citep{neubig_zeno_2023}.
Taken together, LLMs are surpassing NMT models in terms of versatility.

Recent studies have demonstrated that fine-tuning LLMs on parallel data further improves their translations as measured by metrics that reflect overall quality (such as COMET) \citep{li_eliciting_2023,yang_bigtranslate_2023,zeng_tim_2023}.
However, relying on general translation quality metrics and generic test sets does not fully capture the nuanced abilities of LLMs in machine translation.
This oversight raises questions about the retention of LLM-specific advantages --- such as controllability, document-level translation proficiency, and the production of less literal translations --- after fine-tuning on parallel data. 
While it is clear that general machine translation quality improves through fine-tuning, there is a risk that LLMs lose their unique strengths due to catastrophic forgetting \citep{mccloskey_catastrophic_1989,ratcliff_connectionist_1990,luo_empirical_2023}.
Determining the extent of this risk and comparing the effect of various fine-tuning strategies in preserving the qualitative benefits of LLMs remains an important yet unresolved question.

We investigate how qualitative advantages of LLMs change when fine-tuning on parallel data.
We consider LLaMA and Falcon models, with parameter counts ranging from 7 billion up to 65 billion.
The LLM properties we investigate are general translation quality, formality steerability, non-literalness in idiom translations, performance on specialized domains, and performance on document-level input which requires contextualisation of ambiguous tokens.
We compare two fine-tuning strategies for varying data sizes (89K up to 1.4M)  in six translation directions. Our main findings and contributions are:
\begin{itemize}
    \item We show that while fine-tuning LLMs on parallel data enhances overall translation quality as measured by COMET, it simultaneously leads to a decline in important attributes.
    Even when only using 18k fine-tuning samples we observe degradations in formality steering, technical translation through few-shot examples, and contextualization capabilities required for document-level translation.  In general, we find that using larger data sets for fine-tuning data results in more severe degradations, and these trends are consistent across all tested model scales and architectures.
    The exception we observe is in the ability to produce less literal translations, which improves in fine-tuning.
    \item We show that incorporating a mix of monolingual and parallel data during fine-tuning can preserve abilities of LLMs.
    Overall translation quality is enhanced to a greater extent compared to fine-tuning on parallel data alone.    
    \item We introduce a novel evaluation dataset, IdiomsInCtx-MT\footnote{We release the dataset at \url{https://github.com/amazon-science/idioms-incontext-mt}}, to measure non-literalness performance.
    To our knowledge, it is the first dataset that consists of idiomatic expressions in context and their human-written translations.
    It covers 2 language pairs with 3 translation directions.
\end{itemize}
Our findings highlight the importance of creating fine-tuning approaches that enhance general translation quality while also preserving the distinctive capabilities of LLMs for machine translation.

\section{Related Work}

\paragraph{Advantages of LLMs for MT}
Several studies have investigated the use of LLMs for translation.
Generally, current LLMs show strong performance for most language pairs, but lag behind NMT systems when translating into low-resource languages \citep{zhu_multilingual_2023,stap_chatgpt_2023,robinson_chatgpt_2023,kocmi_findings_2023}.
In addition to strong performance, LLMs exhibit certain abilities that are relevant for translation.
NMT systems show a bias towards generating text that is over-represented in the data, such as language varieties \citep{riley_frmt_2023} and formality \citep{rippeth_controlling_2022}, whereas LLMs can easily be controlled for this bias using examples \citep{garcia_unreasonable_2023}.
In addition, examples can be supplied to improve general LLM translation quality via in-context learning \citep{agrawal_-context_2022,moslem_adaptive_2023}.
NMT models are often unable to translate idioms accurately and generate literal translations \citep{dankers_can_2022}.
LLMs produce less literal outputs compared to NMT models, particularly for sentences that contain idiomatic expressions \citep{vilar_prompting_2023,raunak_gpts_2023}.
NMT models are trained on sentence level, and thus do not take into account document context.
LLMs outperform NMT models for document translation in general domains (such as news and social media) \citep{wang_document-level_2023}, as well as in more specialised domains such as literature \citep{karpinska_large_2023}.

\paragraph{Finetuning LLMs for MT}
There are multiple strategies for fine-tuning LLMs for machine translation.
One approach makes use of either a small set of high-quality human-written translations or a set of translation instructions for fine-tuning \citep{li_eliciting_2023,zeng_tim_2023,jiao_parrot_2023,wu2024far,zhu2024fine}.
Another line of work makes use of more traditional machine translation data: parallel data from the web, which is orders of magnitude larger compared to what is used in fine-tuning \citep{yang_bigtranslate_2023,alves_steering_2023,zhang_machine_2023,zhu_extrapolating_2023}.
These strategies are focused on improving general machine translation quality, but it remains unclear what happens to other abilities that are relevant for translation.
We investigate the effect of fine-tuning on relevant abilities for translation using publicly available models.

\section{Fine-tuning LLMs on parallel data}

\subsection{Experimental Setup}

\paragraph{Models}
We use the LLaMA \citep{touvron_llama_2023} 7B, 13B, 30B, 65B, and Falcon \citep{almazrouei_falcon_2023} 7B and 40B language models.

\paragraph{Optimisation}
We refer the reader to Appendix \ref{sec:appendix_experimental_setup} for full details on optimisation, hyperparameters, and instruction formats.

\paragraph{Language directions}
We consider the language directions German (de), Russian (ru), and Chinese (zh) into and out of English (en).

\paragraph{Human-written training data}
Following \citet{xu_paradigm_2023}, we use human-written translations from WMT17 to WMT20, resulting in 89K training examples that are evenly distributed across the language directions we consider.

\paragraph{Web-scraped training data}
Additionally we train models on general domain OPUS \citep{tiedemann_parallel_2012} data from the News-Commentary, WikiTitles, and ParaCrawl \citep{banon_paracrawl_2020} corpora.
To ensure that the resulting data is above an acceptable quality threshold we perform data filtering using an internal Quality Estimation (QE) model, which has a similar architecture as COMETKiwi \citep{rei_cometkiwi_2022} and is based on the InfoXLM-Large pretrained multilingual encoder \citep{chi_infoxlm_2021}. We train our sentence-level QE model on a large internal dataset of human annotations for more than 12 languages, where each translation is rated between 1 (completely random) and 6 (perfect translation).

We use a subset of 1.4M sentence pairs of this filtered data that is evenly distributed across language directions for training. We fine-tune LLaMA models up to 30B parameters on this dataset but leave out larger models because of the high computational cost.

\begin{figure*}[ht]
    \centering
    \includegraphics[width=\linewidth]{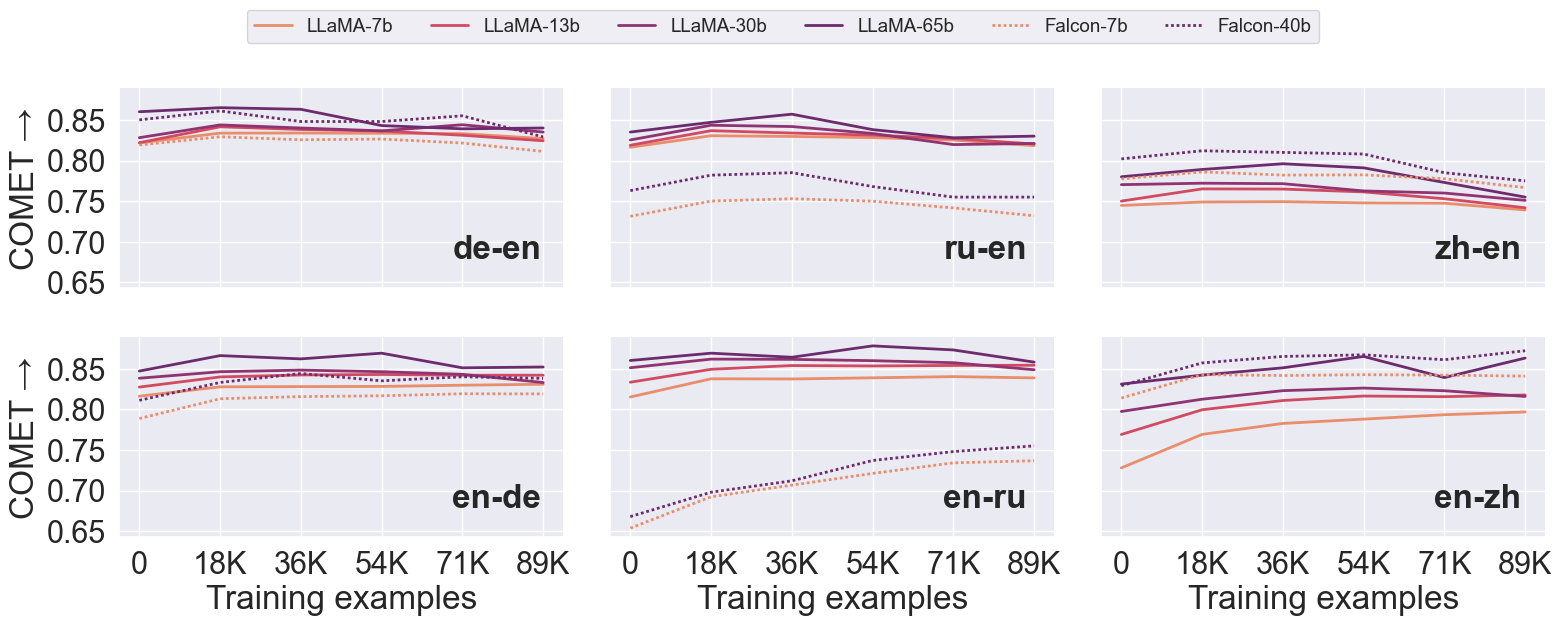}
    \caption{X$\rightarrow$English (top) and English$\rightarrow$X (bottom) COMET scores on WMT22 for models trained on human-written translations with different amounts of training data.
    }
    \label{fig:wmt_general_quality_wmt}
\end{figure*}

\begin{figure}[ht]
    \centering
    \includegraphics[width=\linewidth]{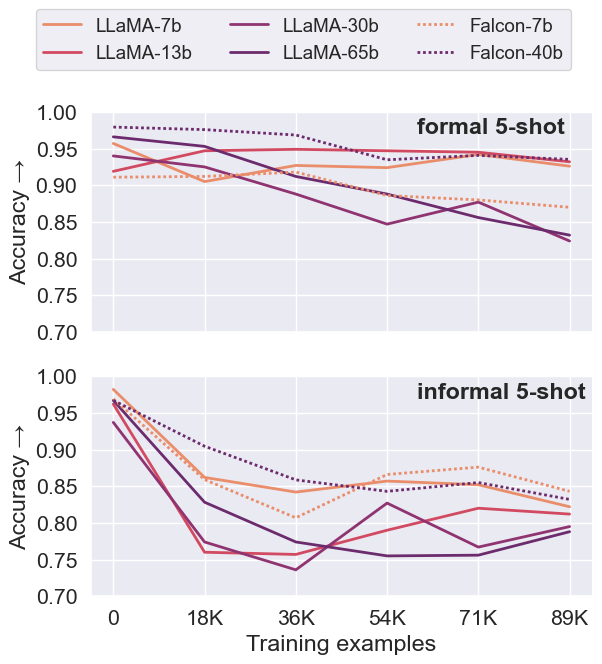}
    \caption{Accuracy of formality markers for models trained on human-written translations. 
    }
    \label{fig:wmt_formality_steering}
\end{figure}

\paragraph{Evaluation data and metrics}
For evaluation, we consider the following test sets:
\begin{itemize}
    \item \textbf{WMT22} To evaluate general machine translation quality, we use WMT22 \citep{kocmi_findings_2022} test sets consisting of news, e-commerce, social, and conversational domains.
    We evaluate all language directions on this test set.
    We evaluate in a 0-shot setting, and report COMET\footnote{model: \texttt{unbabel/wmt22-comet-da}} \citep{rei_comet_2020} scores.

    \item \textbf{CoCoA-MT} To evaluate formality steering ability of LLMs, we make use of the CoCoA-MT \citep{nadejde_cocoa-mt_2022} dataset.
    It consists of 600 test sentences with English source and contrastive target sentences consisting of a formal and informal translation.
    We use German as the target language and report accuracy based on the ratio of correctly predicted formality forms.
    We use 5-shot examples to bias the formality of outputs.\footnote{We also experimented with steering formality through prompting, but the results were inferior to using 5-shot examples.}

    \item \textbf{Law}, \textbf{Medical}, and \textbf{TICO-19} To evaluate the in-context learning ability on technical domains, we consider the Law and Medical test sets from \citet{aharoni_unsupervised_2020}.
    We consider 5-shot inputs.
    We evaluate on German $\leftrightarrow$ English and report COMET scores.
    In addition we evaluate technical abilities on  TICO-19  \citep{anastasopoulos_tico-19_2020},
    which consists of translations in the COVID-19 domain.
    We evaluate on Russian $\leftrightarrow$ English and Chinese $\leftrightarrow$ English.

    \item \textbf{ctxpro} We evaluate performance on longer inputs that includes sentences that require context to be disambiguated correctly by including ctxpro \citep{wicks_identifying_2023}.
    We consider the animacy ambiguity type in German $\rightarrow$ English and Russian $\rightarrow$ English.\footnote{We also experimented with different ambiguity types (auxiliary and gender in out-of-English directions) but the resulting translations are of poor quality, making it impossible to sensibly evaluate decontextualization abilities.}
    While English makes no gender distinction for inanimate objects, some other languages such as Russian do.
    The ambiguous animacy examples in the ctxpro dataset require corresponding document-level context for correct disambiguation.    
    We subsample the test set to 2K examples per language direction.
    The average number of sentences per input is $10.16\pm1.27$.
    We report the generative accuracy score, which measures the accuracy of the contextualization.

    \item \textbf{IdiomsInCtx-MT} We introduce a novel dataset consisting of idiomatic expressions in context and their human-written translations.
    The dataset comprises 2 language pairs: German and Russian paired with English.
    For German, the opposite translation directions are also included.
    Current idiom datasets stem from potentially noisy, web-extracted sources \citep{fadaee-etal-2018-examining}, machine-generated translations \citep{tang_petci_2022}, or are monolingual \citep{haagsma-etal-2020-magpie}.
    In contrast, we use professional translators to create a high-quality evaluation benchmark. 
    Idiomatic expressions and their context sentences were sourced in the respective source language and translated to the target language by professional translators.
    In addition, the dataset contains annotations of the source and target idiomatic expressions for each segment.
    This enables running targeted evaluation on the translations of the idiomatic expressions in addition to general quality metrics.
    We evaluate on English$\rightarrow$German, German$\rightarrow$English and Russian$\rightarrow$English using test splits of 1000 segments.
    We report COMET, LitTER \cite{baziotis-etal-2023-automatic} and MWEScore, an internal multi-reference version of Score$\_$mwe \citep{zaninello-birch-2020-multiword}.
\end{itemize}

\subsection{Results}
\label{sec:parallel_ft_results}

\paragraph{General translation quality improves}
Results on WMT22 for models trained on human-written translations are summarized in Figure~\ref{fig:wmt_general_quality_wmt}.
Consistent with expectations, we observe that fine-tuning generally improves the translation quality, and larger models generally perform better.
Using 89K parallel examples does not always yield better results compared to smaller datasets. 
For most language directions, we notice an initial increase in translation quality, followed by a slight decline.
The most notable increases are found for the out-of-English directions.
In contrast, when models are fine-tuned on a more extensive dataset (up to 1.4 million examples) sourced from the web (refer to Appendix \ref{sec:appendix_additional_results}, Figure \ref{fig:opus_general_quality_wmt}), translation quality continues to improve with the addition of more data.
We hypothesize that this difference stems from the domain-specific composition of the human-written WMT training data, which is exclusively news content.
In contrast, the OPUS dataset is more diverse and includes multiple domains.
This diversity better reflects domain composition of the WMT22 test set, which has content from news, e-commerce, social media and conversational domains.
The improvements are significantly more marked in translations from other languages into English.
A comparative analysis of the best-performing checkpoints of LLaMA models (7B, 13B, 30B) fine-tuned on human-written and OPUS data across 6 translation directions shows a clear trend. In 15 out of 18 cases, models fine-tuned on the larger OPUS dataset achieved superior results (refer to Appendix \ref{sec:appendix_additional_results}, Table \ref{tab:wmt_vs_opus_general_quality}).

\begin{figure*}[ht]
    \centering
    \includegraphics[width=\linewidth]{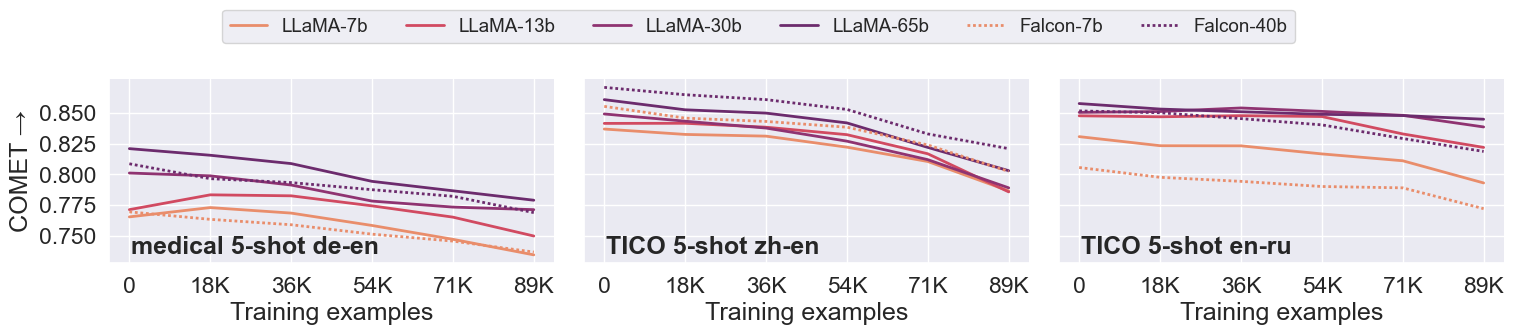}
    \caption{COMET on technical domains using 5-shot examples for models trained on human-written translations. 
    }
    \label{fig:wmt_technical_domains_subset}
\end{figure*}

\begin{figure}[ht]
    \centering
    \includegraphics[width=\linewidth]{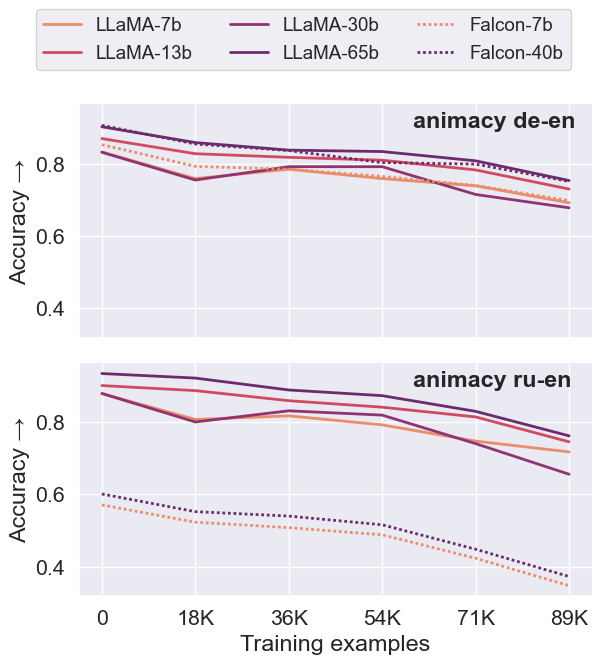}
    \caption{Accuracy of animacy contextualization for German$\rightarrow$English and Russian$\rightarrow$English for models fine-tuned on human-written translations.
    }
    \label{fig:wmt_animacy}
\end{figure}

\paragraph{Formality steering ability degrades}
We show results for formality steering using 5-shot examples for models trained on human-written data in Figure \ref{fig:wmt_formality_steering}.
Notably, the base models exhibit strong performance for this task; for instance, the LLaMA-7b model achieves an accuracy of 0.982 in identifying informal markers.
However, fine-tuning on only 18K examples results in a decline of this ability to 0.862, even though German-English COMET on WMT22 continues to improve up to 36K examples.
The degradation is more pronounced with informal markers, which is likely attributable to the formal bias inherent in the WMT22 training data.
Fine-tuning on more data further degrades formality steering capabilities: there is a significant negative correlation (Spearman's $\rho=-0.46$, $p<0.001$) between formal and informal marker prediction and dataset size.
Notably, these accuracy scores correlate very weakly ($\rho=0.16$, $p<0.001$) with COMET scores, which suggests that comprehensive evaluation of LLMs for machine translation benefits from task-specific metrics.
Further, when examining models trained on OPUS data (see Appendix \ref{sec:appendix_additional_results} Figure \ref{fig:opus_formality_steering}), we find that increasing the amount of fine-tuning data up to 1.4M parallel sentences further exacerbates the degradation.
Again, we observe a significant negative correlation ($\rho=-0.58$, $p<0.001$) between accuracy and dataset size for OPUS-trained models, reinforcing the conclusion that larger parallel datasets during fine-tuning adversely affect formality steering capabilities.

\paragraph{Performance on technical domains degrades}
Next, we evaluate the model capabilities of doing technical translations given 5-shot examples.
Results on human-written training data for a subset of the domains and directions are shown in Figure \ref{fig:wmt_technical_domains_subset}.
In most cases, performance starts to degrade  after only 18K examples.
For example, LLaMA-7b scores 0.8308 COMET on TICO English-Russian, whereas after fine-tuning on 18K examples COMET is 0.8234, a degradation of 0.0074.
Results on the other domains and directions show a similar trend (Appendix~\ref{sec:appendix_additional_results} Figure \ref{fig:wmt_technical_domains_full}).
The COMET scores correlate negatively with datastore size ($\rho=-0.27$, $p<0.001$), indicating that fine-tuning on more data results in larger degradations.
When inspecting models trained on OPUS data, we observe consistent conclusions: few-shot technical domain translation capabilities degrade, and the amount of degradation is dependent on the dataset size (see Appendix \ref{sec:appendix_additional_results} Figure \ref{fig:opus_technical_domains_full}).

\paragraph{Contextualization ability on document-level input degrades}
Our analysis extends to the animacy contextualization accuracy of document-level input.
We show results for models trained on human-written data in Figure \ref{fig:wmt_animacy}.
Mirroring the trend observed in formality steering, a clear degradation in contextualization accuracy is noted upon fine-tuning the models on parallel data.
Again, we observe that fine-tuning on only 18K examples results in a decline of this ability, even though general translation quality on WMT continues to improve.
For example, Falcon-40b scores 0.91 before fine-tuning, which degrades to 0.85 after 18K examples.
The decline can be summarized by a negative correlation between accuracy and the size of the dataset used for fine-tuning ($\rho=-0.55$, $p<0.001$), indicating that contextualization abilities further degrade when fine-tuning on more data.
This trend is not exclusive to WMT data.
A similar pattern emerges when analyzing models trained on smaller human-written translation datasets.
As detailed in Appendix \ref{sec:appendix_additional_results}, Figure \ref{fig:opus_animacy}, we observe a consistent reduction in contextualization accuracy with the addition of more training data, again supported by a negative correlation ($\rho=-0.49$, $p<0.001$).

\begin{figure}[ht]
    \centering
    \includegraphics[width=\linewidth]{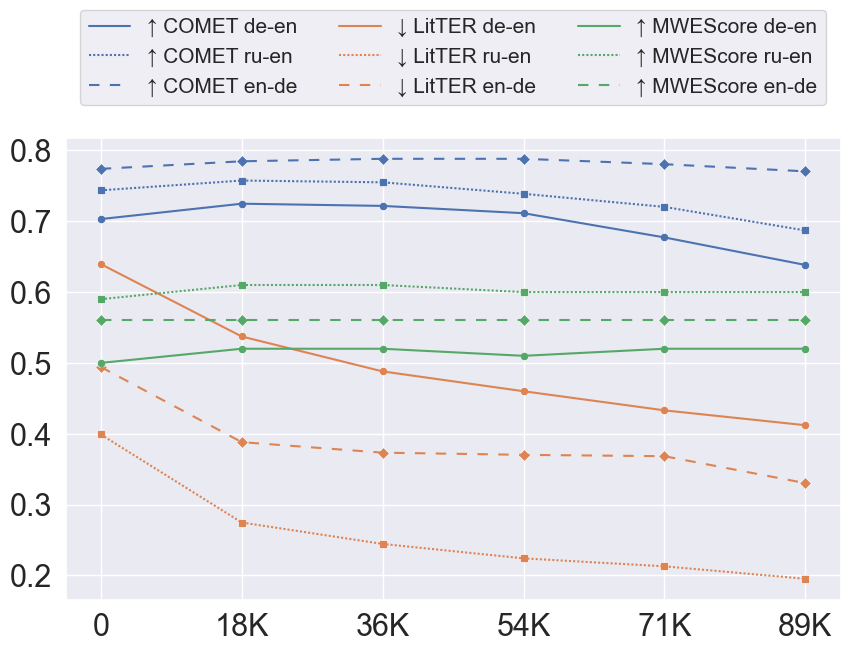}
    \caption{COMET, LitTER and MWEScore on IdiomsInCtx-MT test sets for LLaMA-65b fine-tuned on human-quality parallel data.}
    \label{fig:idioms}
\end{figure}

\paragraph{Performance on idioms remains stable or improves}
We evaluate the quality of idiomatic translations on the IdiomsInCtx-MT test sets.
Since idiomatic translation is inherently difficult, we focus our analysis on the strongest model, LLaMA-65b.
Figure \ref{fig:idioms} shows COMET, LitTER (lower is better) and MWEScore (higher is better) across training checkpoints for 1 epoch.
While LitTER uses input-specific block lists to assess the literalness of translation outputs based on annotations of idiomatic expressions in the input, MWEScore additionally relies on one or more gold translations of those idiomatic expressions and computes a score based on edit distance of output vs reference idiom tokens. 

Similar to the WMT22 test set, we see that COMET scores improve until the first 1-3 checkpoints before stabilizing or starting to decrease.
However, for idiomatic expressions even the targeted metrics LitTER  and MWEScore improve during fine-tuning or least remain stable.
This indicates that even a large open-source model like LLaMA-65b still has room for improvement when it comes to idiomatic translations.
Future work could investigate translation literalness and idiomaticity of LLM translations on stronger models such as GPT-3.5 during fine-tuning.\footnote{https://openai.com/blog/gpt-3-5-turbo-fine-tuning-and-api-updates}

\begin{table*}[t]
    \begin{center}
        \small
        \begin{tabular}{lrrrrrrr}
        model       & de-en     & ru-en     & zh-en     & en-de     & en-ru     & en-zh     & avg       \\\toprule
        base        & $0.8217$  & $0.8163$  & $0.7477$  & $0.8161$  & $0.8151$  & $0.7279$  & $0.7903$  \\
        FT 1:0      & $0.8342$  & $0.8249$  & $0.7435$  & $0.8381$  & $0.8378$  & $0.7771$  & $0.8093$ \\
        FT 2:0      & $0.8360$  & $0.8277$  & $0.7472$  & $\mathbf{0.8400}$  & $0.8436$  & $0.7922$ & $0.8145$ \\\midrule
        FT 1:1      & $\mathbf{0.8380}$  & $\mathbf{0.8280}$  & $\mathbf{0.7519}$  & $0.8375$  & $\mathbf{0.8484}$  & $\mathbf{0.8053}$ & $\mathbf{0.8182}$ \\
        \bottomrule
        \end{tabular}
    \end{center}
    \caption{COMET scores on WMT22. Best scores in \textbf{bold}. Including both parallel and monolingual data during fine-tuning (FT 1:1) results in better translation performance compared to parallel-only fine-tuning (FT 1:0, FT 2:0).}
    \label{tab:mono_results_general}
\end{table*}

\begin{figure*}[ht]
    \centering
    \begin{subfigure}[b]{0.49\linewidth}
        \includegraphics[width=\linewidth]{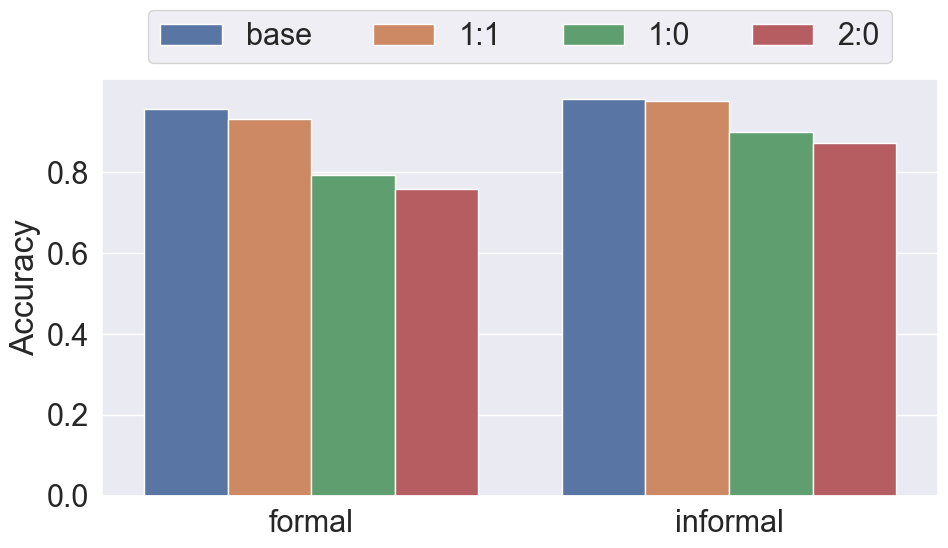}
        \caption{Formality steering}
        \label{fig:sub1}
    \end{subfigure}
    \hfill
    \begin{subfigure}[b]{0.49\linewidth}
        \includegraphics[width=\linewidth]{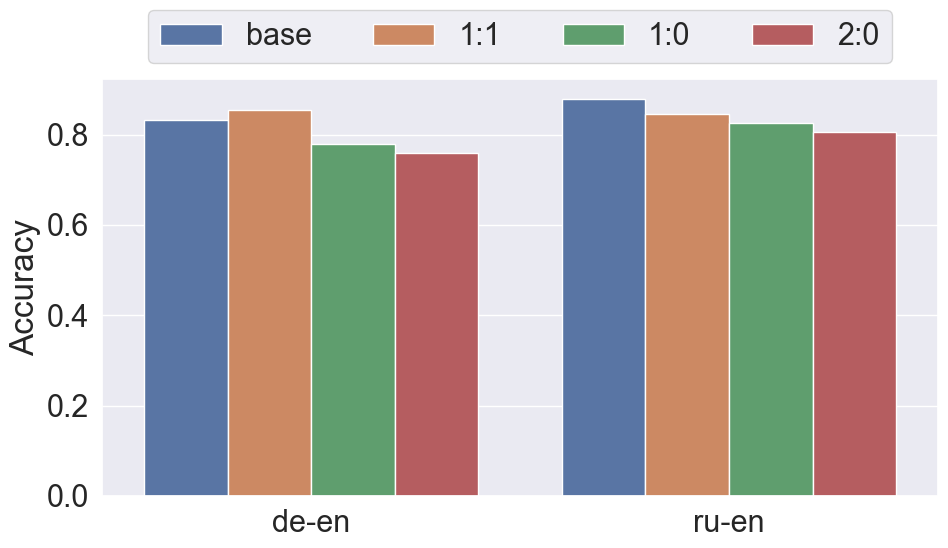}
        \caption{Contextualization}
        \label{fig:sub2}
    \end{subfigure}
    \par\bigskip
    \begin{subfigure}[b]{0.49\linewidth} 
        \includegraphics[width=\linewidth]{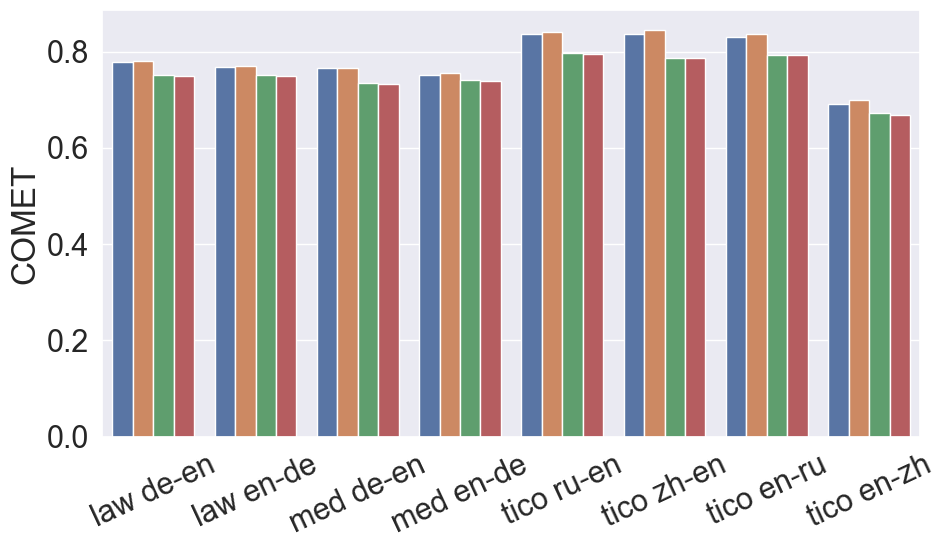}
        \caption{Technical translations}
        \label{fig:sub3}
    \end{subfigure}
    \hfill
    \begin{subfigure}[b]{0.49\linewidth}
        \includegraphics[width=\linewidth]{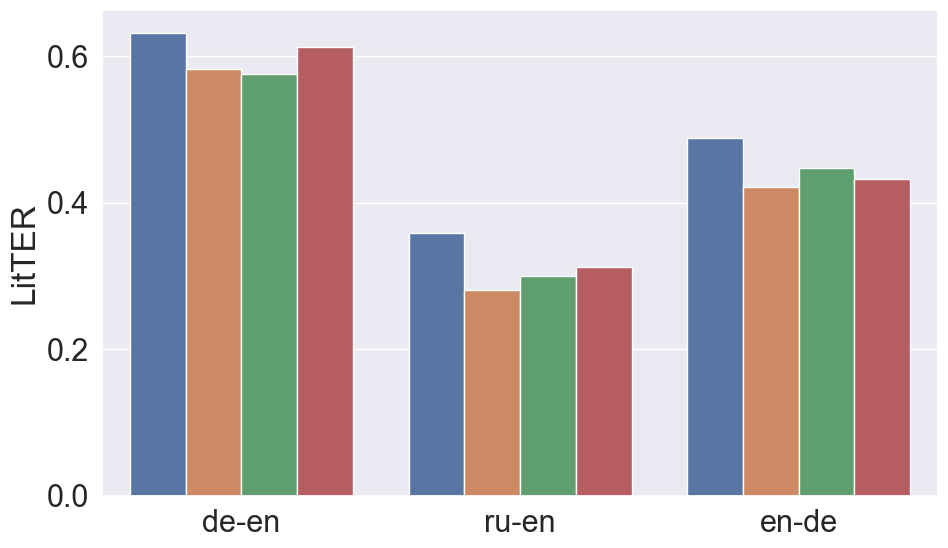}
        \caption{Literalness}
        \label{fig:sub4}
    \end{subfigure}
    \caption{Formality steering accuracy scores (a), Animacy contextualization accuracy of document-level input (b), COMET for few-shot technical translations (c), and LitTER scores for idioms. Base is LLaMA-7b before fine-tuning, 1:1 is fine-tuned on 89K parallel and 89K monolingual data, 1:0 on 89K parallel data, and 2:0 on 178K parallel data. Integrating monolingual data generally results in more preservation of capabilities.}
    \label{fig:mono_results_qualitative}
\end{figure*}

\section{Fine-tuning on a mix of monolingual and parallel data}
Having established that fine-tuning on parallel data leads to a decline in various advantages of LLMs for machine translation, this section delves into strategies to mitigate this degradation.

A potential approach to counteract degradation involves incorporating examples of desired behaviors during fine-tuning.
For instance, the degradation of few-shot capabilities for domain adaptation can be partially mitigated by including few-shot examples during fine-tuning \citep{alves_steering_2023, moslem_fine-tuning_2023}.
However, our aim is to establish a more general solution that prevents degradation across a broader spectrum of behaviors, without the need to specifically include data for each behavior during fine-tuning.
To achieve this, our experiments involve a blend of monolingual and parallel data in the fine-tuning phase.
This strategy stems from the understanding that the pre-training on monolingual data contributed to these beneficial phenomena, and our goal is to retain these qualities during fine-tuning.

\subsection{Experimental setup}
To construct our monolingual dataset, we use the News-Commentary data from WMT22.
This dataset includes document-level information, which we preserve by concatenating sentences within each paragraph to form a single data entry.
The resulting processed data closely resembles the type of data used for LLM pre-training.
We then merged this monolingual dataset with parallel data sourced from OPUS (we sample 89K parallel examples), maintaining a 1:1 ratio and resulting in a total of 178K examples.
We refer to this arrangement as the FT 1:1 setup.
We compare this setup to several baselines: 1) the base model prior to any fine-tuning; 2) the FT 1:0 setup, which involves fine-tuning exclusively on parallel OPUS data (89K); and 3) the FT 2:0 setup, where fine-tuning is conducted on parallel OPUS data equal in size to our mixed monolingual and parallel dataset, totalling 178K examples.
We use LLaMA-7B with a context window size of 2048 tokens for this experiment.
\subsection{Results}

\paragraph{General translation quality further improves} The results, as detailed in Table~\ref{tab:mono_results_general}, show the comparative performance of the baselines and the integration of monolingual data during the fine-tuning phase on general translation quality (WMT22) as measured by COMET.
Including monolingual data (FT 1:1) leads to translations that generally surpass those produced by parallel-only fine-tuning approaches (FT 1:0 and FT 2:0).
Notably, the most significant improvement is observed in the en-zh direction, where the FT 1:1 setup yields an increase of 0.0131 COMET over the best baseline (FT 2:0).
This can be attributed to the pre-training of LLaMA on an English-centric corpus, which contains only minimal amounts of (accidental) Chinese data.
As suggested by \citet{xu_paradigm_2023}, out-of-English capabilities of the model can be substantially augmented through an additional monolingual fine-tuning step, a methodology akin to our approach.

While the enhancement of general translation quality is a beneficial outcome, our primary interest lies in evaluating the ability of our method to preserve and possibly enhance the qualitative behaviors inherent in Large Language Models. This aspect forms the next focus of our investigation.
Figure~\ref{fig:mono_results_qualitative} shows a comparison between our fine-tuning method on formality steering, document-level contextualization, technical translation, and idiom translation tasks.
A consistent trend is observed: the integration of monolingual data with parallel data during fine-tuning generally results in more effective preservation of various translation capabilities.

\paragraph{Minimal formality steering degradation}
Baselines using only parallel data show accuracy drops as great as 0.198 for formal and 0.11 for informal steering.
The inclusion of monolingual data mitigates this degradation, reducing it to just 0.025 for formal and 0.007 for informal steering, albeit some degradation persists when compared to the baseline.

\paragraph{Degradation of contextualization abilities are lessened}
The impact of combining parallel and monolingual data during fine-tuning is also evident here.
For translations from German to English and Russian to English, the loss in accuracy is up to 0.073 and 0.075, respectively, when using only parallel data.
Incorporating monolingual data diminishes this degradation for Russian to English translations (-0.035), and even shows a notable improvement of +0.021 over the base model.

\paragraph{Technical domain performance is enhanced}
The inclusion of monolingual data during fine-tuning enhances performance for English into Russian and Chinese, and vice versa.
For instance, in the TICO English to Chinese translation task, the blended approach of monolingual and parallel data in fine-tuning yields a +0.0089 COMET score improvement over the baseline.
Conversely, relying solely on parallel data results in a 0.022 COMET score decrease (FT 2:0), indicating a substantial differential of 0.03.
For English in and out of German in both the Law and Medical domain, the differences between fine-tuning with monolingual data plus parallel data and the base model are minimal.
In contrast to using parallel data only, we observe no decline.
Notably, we use parallel data up to 178K (FT 2:0), where degradation is relatively modest in the case of few-shot technical domain translation.
When doing extended fine-tuning, this capability will further degrade, as we show in Section \ref{sec:parallel_ft_results}.

\paragraph{Performance on idioms improves}
Including monolingual data (1:1) improves the literalness of translations as measured by LitTER for ru-en and en-de.
However, LLaMA-7b does not demonstrate good performance on idiomatic translations, making it an easy baseline to improve upon.

These findings underscore the nuanced benefits of incorporating monolingual data when fine-tuning English-centric LLMs for machine translation tasks, specifically in preserving task-relevant capabilities.
However, for long-term progress, we advocate for the development of LLMs with multilingual data in mind.
In this approach, parallel data would be combined with monolingual data during the pre-training phase \citep{anil_palm_2023,wei_polylm_2023}.
Nevertheless, even when beginning with a more robust multilingual LLM for translation purposes, the exploration of fine-tuning strategies that preserve the emerged capabilities of LLMs remains critical, especially when adapting these models for various use cases and objectives.

\section{Conclusion}
We investigated how fine-tuning on parallel data affects the qualitative advantages of LLMs for machine translation.
While previous research predominantly focused on summary quality metrics like COMET, our findings reveal a more complex interplay between fine-tuning and LLM capabilities.
Consistent with prior work, we find that fine-tuning enhances the general translation quality of LLMs.
However, we show that fine-tuning adversely impacts several important qualitative advantages of LLMs.
We observe declines in the abilities of LLMs to 1) perform formality steering, 2) perform technical translation through few-shot examples, as well as 3) a decrease in their document-level translation capabilities.
The ability to produce non-literal translations shows improvement post fine-tuning, likely because the publicly available LLMs we investigate do not perform strongly on this task to begin with.
Furthermore, our results indicate that these degradations are more pronounced for larger fine-tuning datasets, even when generic translation quality continues to improve.
These trends are consistent across different model scales (7b up to 65b), underscoring the generalizability of our findings.
To prevent these degradations, we develop a fine-tuning method tailored for machine translation, that combines monolingual and parallel data.
We show that this approach mitigates the degradation of LLMs' qualitative advantages, thereby preserving their capabilities while improving general translation quality.

\section*{Limitations}
Because of the high cost of repeatedly fine-tuning LLMs of different sizes, we limited ourselves to 6 translation directions (German, Chinese, and Russian in and out of English). The impact of fine-tuning on emergent abilities of LLMs when translating in and out of low-resource languages is not studied in our work.

\section*{Ethics statement}
The IdiomsInCtx-MT dataset is annotated by professional translators and they were all paid a fair rate.

\subsubsection*{Acknowledgments}
We thank Zach Hille and Arda Keskiner for their support in our experimental work.

Christof Monz acknowledges funding by the Netherlands Organization for Scientific Research (NWO) under project number VI.C.192.080.

\bibliography{anthology,references, custom}

\begin{thebibliography}{55}
\expandafter\ifx\csname natexlab\endcsname\relax\def\natexlab#1{#1}\fi

\bibitem[{Agrawal et~al.(2022)Agrawal, Zhou, Lewis, Zettlemoyer, and Ghazvininejad}]{agrawal_-context_2022}
Sweta Agrawal, Chunting Zhou, Mike Lewis, Luke Zettlemoyer, and Marjan Ghazvininejad. 2022.
\newblock \href {http://arxiv.org/abs/2212.02437} {In-context {Examples} {Selection} for {Machine} {Translation}}.
\newblock ArXiv:2212.02437 [cs].

\bibitem[{Agrawal et~al.(2023)Agrawal, Zhou, Lewis, Zettlemoyer, and Ghazvininejad}]{agrawal_-context_2023}
Sweta Agrawal, Chunting Zhou, Mike Lewis, Luke Zettlemoyer, and Marjan Ghazvininejad. 2023.
\newblock \href {https://doi.org/10.18653/v1/2023.findings-acl.564} {In-context {Examples} {Selection} for {Machine} {Translation}}.
\newblock In \emph{Findings of the {Association} for {Computational} {Linguistics}: {ACL} 2023}, pages 8857--8873, Toronto, Canada. Association for Computational Linguistics.

\bibitem[{Aharoni and Goldberg(2020)}]{aharoni_unsupervised_2020}
Roee Aharoni and Yoav Goldberg. 2020.
\newblock \href {https://doi.org/10.18653/v1/2020.acl-main.692} {Unsupervised {Domain} {Clusters} in {Pretrained} {Language} {Models}}.
\newblock In \emph{Proceedings of the 58th {Annual} {Meeting} of the {Association} for {Computational} {Linguistics}}, pages 7747--7763, Online. Association for Computational Linguistics.

\bibitem[{Almazrouei et~al.(2023)Almazrouei, Alobeidli, Alshamsi, Cappelli, Cojocaru, Debbah, Goffinet, Hesslow, Launay, Malartic, Mazzotta, Noune, Pannier, and Penedo}]{almazrouei_falcon_2023}
Ebtesam Almazrouei, Hamza Alobeidli, Abdulaziz Alshamsi, Alessandro Cappelli, Ruxandra Cojocaru, Mérouane Debbah, Étienne Goffinet, Daniel Hesslow, Julien Launay, Quentin Malartic, Daniele Mazzotta, Badreddine Noune, Baptiste Pannier, and Guilherme Penedo. 2023.
\newblock \href {http://arxiv.org/abs/2311.16867} {The {Falcon} {Series} of {Open} {Language} {Models}}.
\newblock ArXiv:2311.16867 [cs].

\bibitem[{Alves et~al.(2023)Alves, Guerreiro, Alves, Pombal, Rei, de~Souza, Colombo, and Martins}]{alves_steering_2023}
Duarte~M. Alves, Nuno~M. Guerreiro, João Alves, José Pombal, Ricardo Rei, José G.~C. de~Souza, Pierre Colombo, and André F.~T. Martins. 2023.
\newblock \href {http://arxiv.org/abs/2310.13448} {Steering {Large} {Language} {Models} for {Machine} {Translation} with {Finetuning} and {In}-{Context} {Learning}}.
\newblock ArXiv:2310.13448 [cs].

\bibitem[{Anastasopoulos et~al.(2020)Anastasopoulos, Cattelan, Dou, Federico, Federmann, Genzel, Guzmán, Hu, Hughes, Koehn, Lazar, Lewis, Neubig, Niu, Öktem, Paquin, Tang, and Tur}]{anastasopoulos_tico-19_2020}
Antonios Anastasopoulos, Alessandro Cattelan, Zi-Yi Dou, Marcello Federico, Christian Federmann, Dmitriy Genzel, Franscisco Guzmán, Junjie Hu, Macduff Hughes, Philipp Koehn, Rosie Lazar, Will Lewis, Graham Neubig, Mengmeng Niu, Alp Öktem, Eric Paquin, Grace Tang, and Sylwia Tur. 2020.
\newblock \href {https://doi.org/10.18653/v1/2020.nlpcovid19-2.5} {{TICO}-19: the {Translation} {Initiative} for {COvid}-19}.
\newblock In \emph{Proceedings of the 1st {Workshop} on {NLP} for {COVID}-19 ({Part} 2) at {EMNLP} 2020}, Online. Association for Computational Linguistics.

\bibitem[{Anil et~al.(2023)Anil, Dai, Firat, Johnson, Lepikhin, Passos, Shakeri, Taropa, Bailey, Chen, Chu, Clark, Shafey, Huang, Meier-Hellstern, Mishra, Moreira, Omernick, Robinson, Ruder, Tay, Xiao, Xu, Zhang, Abrego, Ahn, Austin, Barham, Botha, Bradbury, Brahma, Brooks, Catasta, Cheng, Cherry, Choquette-Choo, Chowdhery, Crepy, Dave, Dehghani, Dev, Devlin, Díaz, Du, Dyer, Feinberg, Feng, Fienber, Freitag, Garcia, Gehrmann, Gonzalez, Gur-Ari, Hand, Hashemi, Hou, Howland, Hu, Hui, Hurwitz, Isard, Ittycheriah, Jagielski, Jia, Kenealy, Krikun, Kudugunta, Lan, Lee, Lee, Li, Li, Li, Li, Li, Lim, Lin, Liu, Liu, Maggioni, Mahendru, Maynez, Misra, Moussalem, Nado, Nham, Ni, Nystrom, Parrish, Pellat, Polacek, Polozov, Pope, Qiao, Reif, Richter, Riley, Ros, Roy, Saeta, Samuel, Shelby, Slone, Smilkov, So, Sohn, Tokumine, Valter, Vasudevan, Vodrahalli, Wang, Wang, Wang, Wang, Wieting, Wu, Xu, Xu, Xue, Yin, Yu, Zhang, Zheng, Zheng, Zhou, Zhou, Petrov, and Wu}]{anil_palm_2023}
Rohan Anil, Andrew~M. Dai, Orhan Firat, Melvin Johnson, Dmitry Lepikhin, Alexandre Passos, Siamak Shakeri, Emanuel Taropa, Paige Bailey, Zhifeng Chen, Eric Chu, Jonathan~H. Clark, Laurent~El Shafey, Yanping Huang, Kathy Meier-Hellstern, Gaurav Mishra, Erica Moreira, Mark Omernick, Kevin Robinson, Sebastian Ruder, Yi~Tay, Kefan Xiao, Yuanzhong Xu, Yujing Zhang, Gustavo~Hernandez Abrego, Junwhan Ahn, Jacob Austin, Paul Barham, Jan Botha, James Bradbury, Siddhartha Brahma, Kevin Brooks, Michele Catasta, Yong Cheng, Colin Cherry, Christopher~A. Choquette-Choo, Aakanksha Chowdhery, Clément Crepy, Shachi Dave, Mostafa Dehghani, Sunipa Dev, Jacob Devlin, Mark Díaz, Nan Du, Ethan Dyer, Vlad Feinberg, Fangxiaoyu Feng, Vlad Fienber, Markus Freitag, Xavier Garcia, Sebastian Gehrmann, Lucas Gonzalez, Guy Gur-Ari, Steven Hand, Hadi Hashemi, Le~Hou, Joshua Howland, Andrea Hu, Jeffrey Hui, Jeremy Hurwitz, Michael Isard, Abe Ittycheriah, Matthew Jagielski, Wenhao Jia, Kathleen Kenealy, Maxim Krikun, Sneha Kudugunta, Chang
  Lan, Katherine Lee, Benjamin Lee, Eric Li, Music Li, Wei Li, YaGuang Li, Jian Li, Hyeontaek Lim, Hanzhao Lin, Zhongtao Liu, Frederick Liu, Marcello Maggioni, Aroma Mahendru, Joshua Maynez, Vedant Misra, Maysam Moussalem, Zachary Nado, John Nham, Eric Ni, Andrew Nystrom, Alicia Parrish, Marie Pellat, Martin Polacek, Alex Polozov, Reiner Pope, Siyuan Qiao, Emily Reif, Bryan Richter, Parker Riley, Alex~Castro Ros, Aurko Roy, Brennan Saeta, Rajkumar Samuel, Renee Shelby, Ambrose Slone, Daniel Smilkov, David~R. So, Daniel Sohn, Simon Tokumine, Dasha Valter, Vijay Vasudevan, Kiran Vodrahalli, Xuezhi Wang, Pidong Wang, Zirui Wang, Tao Wang, John Wieting, Yuhuai Wu, Kelvin Xu, Yunhan Xu, Linting Xue, Pengcheng Yin, Jiahui Yu, Qiao Zhang, Steven Zheng, Ce~Zheng, Weikang Zhou, Denny Zhou, Slav Petrov, and Yonghui Wu. 2023.
\newblock \href {http://arxiv.org/abs/2305.10403} {{PaLM} 2 {Technical} {Report}}.
\newblock ArXiv:2305.10403 [cs].

\bibitem[{Baziotis et~al.(2023)Baziotis, Mathur, and Hasler}]{baziotis-etal-2023-automatic}
Christos Baziotis, Prashant Mathur, and Eva Hasler. 2023.
\newblock \href {https://doi.org/10.18653/v1/2023.eacl-main.267} {Automatic evaluation and analysis of idioms in neural machine translation}.
\newblock In \emph{Proceedings of the 17th Conference of the European Chapter of the Association for Computational Linguistics}, pages 3682--3700, Dubrovnik, Croatia. Association for Computational Linguistics.

\bibitem[{Bañón et~al.(2020)Bañón, Chen, Haddow, Heafield, Hoang, Esplà-Gomis, Forcada, Kamran, Kirefu, Koehn, Ortiz~Rojas, Pla~Sempere, Ramírez-Sánchez, Sarrías, Strelec, Thompson, Waites, Wiggins, and Zaragoza}]{banon_paracrawl_2020}
Marta Bañón, Pinzhen Chen, Barry Haddow, Kenneth Heafield, Hieu Hoang, Miquel Esplà-Gomis, Mikel~L. Forcada, Amir Kamran, Faheem Kirefu, Philipp Koehn, Sergio Ortiz~Rojas, Leopoldo Pla~Sempere, Gema Ramírez-Sánchez, Elsa Sarrías, Marek Strelec, Brian Thompson, William Waites, Dion Wiggins, and Jaume Zaragoza. 2020.
\newblock \href {https://doi.org/10.18653/v1/2020.acl-main.417} {{ParaCrawl}: {Web}-{Scale} {Acquisition} of {Parallel} {Corpora}}.
\newblock In \emph{Proceedings of the 58th {Annual} {Meeting} of the {Association} for {Computational} {Linguistics}}, pages 4555--4567, Online. Association for Computational Linguistics.

\bibitem[{Brown et~al.(2020)Brown, Mann, Ryder, Subbiah, Kaplan, Dhariwal, Neelakantan, Shyam, Sastry, Askell, Agarwal, Herbert-Voss, Krueger, Henighan, Child, Ramesh, Ziegler, Wu, Winter, Hesse, Chen, Sigler, Litwin, Gray, Chess, Clark, Berner, McCandlish, Radford, Sutskever, and Amodei}]{brown_language_2020}
Tom Brown, Benjamin Mann, Nick Ryder, Melanie Subbiah, Jared~D Kaplan, Prafulla Dhariwal, Arvind Neelakantan, Pranav Shyam, Girish Sastry, Amanda Askell, Sandhini Agarwal, Ariel Herbert-Voss, Gretchen Krueger, Tom Henighan, Rewon Child, Aditya Ramesh, Daniel Ziegler, Jeffrey Wu, Clemens Winter, Chris Hesse, Mark Chen, Eric Sigler, Mateusz Litwin, Scott Gray, Benjamin Chess, Jack Clark, Christopher Berner, Sam McCandlish, Alec Radford, Ilya Sutskever, and Dario Amodei. 2020.
\newblock \href {https://proceedings.neurips.cc/paper_files/paper/2020/file/1457c0d6bfcb4967418bfb8ac142f64a-Paper.pdf} {Language {Models} are {Few}-{Shot} {Learners}}.
\newblock In \emph{Advances in {Neural} {Information} {Processing} {Systems}}, volume~33, pages 1877--1901. Curran Associates, Inc.

\bibitem[{Chi et~al.(2021)Chi, Dong, Wei, Yang, Singhal, Wang, Song, Mao, Huang, and Zhou}]{chi_infoxlm_2021}
Zewen Chi, Li~Dong, Furu Wei, Nan Yang, Saksham Singhal, Wenhui Wang, Xia Song, Xian-Ling Mao, Heyan Huang, and Ming Zhou. 2021.
\newblock \href {https://doi.org/10.18653/v1/2021.naacl-main.280} {{InfoXLM}: {An} {Information}-{Theoretic} {Framework} for {Cross}-{Lingual} {Language} {Model} {Pre}-{Training}}.
\newblock In \emph{Proceedings of the 2021 {Conference} of the {North} {American} {Chapter} of the {Association} for {Computational} {Linguistics}: {Human} {Language} {Technologies}}, pages 3576--3588, Online. Association for Computational Linguistics.

\bibitem[{Dankers et~al.(2022)Dankers, Lucas, and Titov}]{dankers_can_2022}
Verna Dankers, Christopher Lucas, and Ivan Titov. 2022.
\newblock \href {https://doi.org/10.18653/v1/2022.acl-long.252} {Can {Transformer} be {Too} {Compositional}? {Analysing} {Idiom} {Processing} in {Neural} {Machine} {Translation}}.
\newblock In \emph{Proceedings of the 60th {Annual} {Meeting} of the {Association} for {Computational} {Linguistics} ({Volume} 1: {Long} {Papers})}, pages 3608--3626, Dublin, Ireland. Association for Computational Linguistics.

\bibitem[{Dettmers et~al.(2023)Dettmers, Pagnoni, Holtzman, and Zettlemoyer}]{dettmers_qlora_2023}
Tim Dettmers, Artidoro Pagnoni, Ari Holtzman, and Luke Zettlemoyer. 2023.
\newblock Qlora: {Efficient} finetuning of quantized llms.
\newblock \emph{arXiv preprint arXiv:2305.14314}.

\bibitem[{Fadaee et~al.(2018)Fadaee, Bisazza, and Monz}]{fadaee-etal-2018-examining}
Marzieh Fadaee, Arianna Bisazza, and Christof Monz. 2018.
\newblock \href {https://aclanthology.org/L18-1148} {Examining the tip of the iceberg: {A} data set for idiom translation}.
\newblock In \emph{Proceedings of the eleventh international conference on language resources and evaluation ({LREC} 2018)}, Miyazaki, Japan. European Language Resources Association (ELRA).

\bibitem[{Garcia et~al.(2023)Garcia, Bansal, Cherry, Foster, Krikun, Feng, Johnson, and Firat}]{garcia_unreasonable_2023}
Xavier Garcia, Yamini Bansal, Colin Cherry, George Foster, Maxim Krikun, Fangxiaoyu Feng, Melvin Johnson, and Orhan Firat. 2023.
\newblock \href {http://arxiv.org/abs/2302.01398} {The unreasonable effectiveness of few-shot learning for machine translation}.
\newblock ArXiv:2302.01398 [cs].

\bibitem[{Haagsma et~al.(2020)Haagsma, Bos, and Nissim}]{haagsma-etal-2020-magpie}
Hessel Haagsma, Johan Bos, and Malvina Nissim. 2020.
\newblock \href {https://aclanthology.org/2020.lrec-1.35} {{MAGPIE}: {A} large corpus of potentially idiomatic expressions}.
\newblock In \emph{Proceedings of the twelfth language resources and evaluation conference}, pages 279--287, Marseille, France. European Language Resources Association.

\bibitem[{Jiao et~al.(2023)Jiao, Huang, Wang, Wang, Shi, and Tu}]{jiao_parrot_2023}
Wenxiang Jiao, Jen-tse Huang, Wenxuan Wang, Xing Wang, Shuming Shi, and Zhaopeng Tu. 2023.
\newblock \href {http://arxiv.org/abs/2304.02426} {{ParroT}: {Translating} {During} {Chat} {Using} {Large} {Language} {Models}}.
\newblock ArXiv:2304.02426 [cs].

\bibitem[{Johnson et~al.(2019)Johnson, Douze, and Jégou}]{johnson_billion-scale_2019}
Jeff Johnson, Matthijs Douze, and Hervé Jégou. 2019.
\newblock Billion-scale similarity search with {GPUs}.
\newblock \emph{IEEE Transactions on Big Data}, 7(3):535--547.
\newblock Publisher: IEEE.

\bibitem[{Karpinska and Iyyer(2023)}]{karpinska_large_2023}
Marzena Karpinska and Mohit Iyyer. 2023.
\newblock \href {http://arxiv.org/abs/2304.03245} {Large language models effectively leverage document-level context for literary translation, but critical errors persist}.
\newblock ArXiv:2304.03245 [cs].

\bibitem[{Kocmi et~al.(2023)Kocmi, Avramidis, Bawden, Bojar, Dvorkovich, Federmann, Fishel, Freitag, Gowda, Grundkiewicz, Haddow, Koehn, Marie, Monz, Morishita, Murray, Nagata, Nakazawa, Popel, Popović, and Shmatova}]{kocmi_findings_2023}
Tom Kocmi, Eleftherios Avramidis, Rachel Bawden, Ondřej Bojar, Anton Dvorkovich, Christian Federmann, Mark Fishel, Markus Freitag, Thamme Gowda, Roman Grundkiewicz, Barry Haddow, Philipp Koehn, Benjamin Marie, Christof Monz, Makoto Morishita, Kenton Murray, Makoto Nagata, Toshiaki Nakazawa, Martin Popel, Maja Popović, and Mariya Shmatova. 2023.
\newblock \href {https://doi.org/10.18653/v1/2023.wmt-1.1} {Findings of the 2023 {Conference} on {Machine} {Translation} ({WMT23}): {LLMs} {Are} {Here} but {Not} {Quite} {There} {Yet}}.
\newblock In \emph{Proceedings of the {Eighth} {Conference} on {Machine} {Translation}}, pages 1--42, Singapore. Association for Computational Linguistics.

\bibitem[{Kocmi et~al.(2022)Kocmi, Bawden, Bojar, Dvorkovich, Federmann, Fishel, Gowda, Graham, Grundkiewicz, Haddow, Knowles, Koehn, Monz, Morishita, Nagata, Nakazawa, Novák, Popel, and Popović}]{kocmi_findings_2022}
Tom Kocmi, Rachel Bawden, Ondřej Bojar, Anton Dvorkovich, Christian Federmann, Mark Fishel, Thamme Gowda, Yvette Graham, Roman Grundkiewicz, Barry Haddow, Rebecca Knowles, Philipp Koehn, Christof Monz, Makoto Morishita, Masaaki Nagata, Toshiaki Nakazawa, Michal Novák, Martin Popel, and Maja Popović. 2022.
\newblock \href {https://aclanthology.org/2022.wmt-1.1} {Findings of the 2022 {Conference} on {Machine} {Translation} ({WMT22})}.
\newblock In \emph{Proceedings of the {Seventh} {Conference} on {Machine} {Translation} ({WMT})}, pages 1--45, Abu Dhabi, United Arab Emirates (Hybrid). Association for Computational Linguistics.

\bibitem[{Li et~al.(2023)Li, Zhou, Huang, Cheng, and Chen}]{li_eliciting_2023}
Jiahuan Li, Hao Zhou, Shujian Huang, Shanbo Cheng, and Jiajun Chen. 2023.
\newblock \href {http://arxiv.org/abs/2305.15083} {Eliciting the {Translation} {Ability} of {Large} {Language} {Models} via {Multilingual} {Finetuning} with {Translation} {Instructions}}.
\newblock ArXiv:2305.15083 [cs].

\bibitem[{Luo et~al.(2023)Luo, Yang, Meng, Li, Zhou, and Zhang}]{luo_empirical_2023}
Yun Luo, Zhen Yang, Fandong Meng, Yafu Li, Jie Zhou, and Yue Zhang. 2023.
\newblock \href {https://doi.org/10.48550/arXiv.2308.08747} {An {Empirical} {Study} of {Catastrophic} {Forgetting} in {Large} {Language} {Models} {During} {Continual} {Fine}-tuning}.
\newblock ArXiv:2308.08747 [cs].

\bibitem[{McCloskey and Cohen(1989)}]{mccloskey_catastrophic_1989}
Michael McCloskey and Neal~J Cohen. 1989.
\newblock Catastrophic interference in connectionist networks: {The} sequential learning problem.
\newblock In \emph{Psychology of learning and motivation}, volume~24, pages 109--165. Elsevier.

\bibitem[{Moslem et~al.(2023{\natexlab{a}})Moslem, Haque, Kelleher, and Way}]{moslem_adaptive_2023}
Yasmin Moslem, Rejwanul Haque, John~D. Kelleher, and Andy Way. 2023{\natexlab{a}}.
\newblock \href {http://arxiv.org/abs/2301.13294} {Adaptive {Machine} {Translation} with {Large} {Language} {Models}}.
\newblock ArXiv:2301.13294 [cs].

\bibitem[{Moslem et~al.(2023{\natexlab{b}})Moslem, Haque, and Way}]{moslem_fine-tuning_2023}
Yasmin Moslem, Rejwanul Haque, and Andy Way. 2023{\natexlab{b}}.
\newblock \href {http://arxiv.org/abs/2312.12740} {Fine-tuning {Large} {Language} {Models} for {Adaptive} {Machine} {Translation}}.
\newblock ArXiv:2312.12740 [cs].

\bibitem[{Nadejde et~al.(2022)Nadejde, Currey, Hsu, Niu, Federico, and Dinu}]{nadejde_cocoa-mt_2022}
Maria Nadejde, Anna Currey, Benjamin Hsu, Xing Niu, Marcello Federico, and Georgiana Dinu. 2022.
\newblock \href {https://doi.org/10.18653/v1/2022.findings-naacl.47} {{CoCoA}-{MT}: {A} {Dataset} and {Benchmark} for {Contrastive} {Controlled} {MT} with {Application} to {Formality}}.
\newblock In \emph{Findings of the {Association} for {Computational} {Linguistics}: {NAACL} 2022}, pages 616--632, Seattle, United States. Association for Computational Linguistics.

\bibitem[{Neubig(2023)}]{neubig_zeno_2023}
Graham Neubig. 2023.
\newblock \href {https://github.com/zeno-ml/zeno-build/tree/main/examples/analysis_gpt_mt/report} {Zeno {GPT}-{MT} {Report}}.
\newblock Technical report.

\bibitem[{Rasley et~al.(2020)Rasley, Rajbhandari, Ruwase, and He}]{rasley_deepspeed_2020}
Jeff Rasley, Samyam Rajbhandari, Olatunji Ruwase, and Yuxiong He. 2020.
\newblock Deepspeed: {System} optimizations enable training deep learning models with over 100 billion parameters.
\newblock In \emph{Proceedings of the 26th {ACM} {SIGKDD} {International} {Conference} on {Knowledge} {Discovery} \& {Data} {Mining}}, pages 3505--3506.

\bibitem[{Ratcliff(1990)}]{ratcliff_connectionist_1990}
Roger Ratcliff. 1990.
\newblock Connectionist models of recognition memory: constraints imposed by learning and forgetting functions.
\newblock \emph{Psychological review}, 97(2):285.
\newblock Publisher: American Psychological Association.

\bibitem[{Raunak et~al.(2023)Raunak, Menezes, Post, and Awadalla}]{raunak_gpts_2023}
Vikas Raunak, Arul Menezes, Matt Post, and Hany~Hassan Awadalla. 2023.
\newblock \href {https://aclanthology.org/2023.acl-short.90} {Do {GPTs} {Produce} {Less} {Literal} {Translations}?}
\newblock In \emph{Proceedings of the 61st {Annual} {Meeting} of the {Association} for {Computational} {Linguistics} ({Volume} 2: {Short} {Papers})}, pages 1041--1050, Toronto, Canada. Association for Computational Linguistics.
\newblock ArXiv:2305.16806 [cs].

\bibitem[{Rei et~al.(2020)Rei, Stewart, Farinha, and Lavie}]{rei_comet_2020}
Ricardo Rei, Craig Stewart, Ana~C Farinha, and Alon Lavie. 2020.
\newblock \href {https://doi.org/10.18653/v1/2020.emnlp-main.213} {{COMET}: {A} {Neural} {Framework} for {MT} {Evaluation}}.
\newblock In \emph{Proceedings of the 2020 {Conference} on {Empirical} {Methods} in {Natural} {Language} {Processing} ({EMNLP})}, pages 2685--2702, Online. Association for Computational Linguistics.

\bibitem[{Rei et~al.(2022)Rei, Treviso, Guerreiro, Zerva, Farinha, Maroti, C.~de Souza, Glushkova, Alves, Coheur, Lavie, and Martins}]{rei_cometkiwi_2022}
Ricardo Rei, Marcos Treviso, Nuno~M. Guerreiro, Chrysoula Zerva, Ana~C Farinha, Christine Maroti, José~G. C.~de Souza, Taisiya Glushkova, Duarte Alves, Luisa Coheur, Alon Lavie, and André F.~T. Martins. 2022.
\newblock \href {https://aclanthology.org/2022.wmt-1.60} {{CometKiwi}: {IST}-{Unbabel} 2022 {Submission} for the {Quality} {Estimation} {Shared} {Task}}.
\newblock In \emph{Proceedings of the {Seventh} {Conference} on {Machine} {Translation} ({WMT})}, pages 634--645, Abu Dhabi, United Arab Emirates (Hybrid). Association for Computational Linguistics.

\bibitem[{Reimers and Gurevych(2019)}]{reimers_sentence-bert_2019}
Nils Reimers and Iryna Gurevych. 2019.
\newblock \href {https://doi.org/10.18653/v1/D19-1410} {Sentence-{BERT}: {Sentence} {Embeddings} using {Siamese} {BERT}-{Networks}}.
\newblock In \emph{Proceedings of the 2019 {Conference} on {Empirical} {Methods} in {Natural} {Language} {Processing} and the 9th {International} {Joint} {Conference} on {Natural} {Language} {Processing} ({EMNLP}-{IJCNLP})}, pages 3982--3992, Hong Kong, China. Association for Computational Linguistics.

\bibitem[{Riley et~al.(2023)Riley, Dozat, Botha, Garcia, Garrette, Riesa, Firat, and Constant}]{riley_frmt_2023}
Parker Riley, Timothy Dozat, Jan~A. Botha, Xavier Garcia, Dan Garrette, Jason Riesa, Orhan Firat, and Noah Constant. 2023.
\newblock \href {https://doi.org/10.1162/tacl_a_00568} {{FRMT}: {A} {Benchmark} for {Few}-{Shot} {Region}-{Aware} {Machine} {Translation}}.
\newblock \emph{Transactions of the Association for Computational Linguistics}, 11:671--685.

\bibitem[{Rippeth et~al.(2022)Rippeth, Agrawal, and Carpuat}]{rippeth_controlling_2022}
Elijah Rippeth, Sweta Agrawal, and Marine Carpuat. 2022.
\newblock \href {https://doi.org/10.18653/v1/2022.iwslt-1.30} {Controlling {Translation} {Formality} {Using} {Pre}-trained {Multilingual} {Language} {Models}}.
\newblock In \emph{Proceedings of the 19th {International} {Conference} on {Spoken} {Language} {Translation} ({IWSLT} 2022)}, pages 327--340, Dublin, Ireland (in-person and online). Association for Computational Linguistics.

\bibitem[{Robinson et~al.(2023)Robinson, Ogayo, Mortensen, and Neubig}]{robinson_chatgpt_2023}
Nathaniel Robinson, Perez Ogayo, David~R. Mortensen, and Graham Neubig. 2023.
\newblock \href {https://doi.org/10.18653/v1/2023.wmt-1.40} {{ChatGPT} {MT}: {Competitive} for {High}- (but {Not} {Low}-) {Resource} {Languages}}.
\newblock In \emph{Proceedings of the {Eighth} {Conference} on {Machine} {Translation}}, pages 392--418, Singapore. Association for Computational Linguistics.

\bibitem[{Stap and Araabi(2023)}]{stap_chatgpt_2023}
David Stap and Ali Araabi. 2023.
\newblock \href {https://doi.org/10.18653/v1/2023.americasnlp-1.17} {{ChatGPT} is not a good indigenous translator}.
\newblock In \emph{Proceedings of the {Workshop} on {Natural} {Language} {Processing} for {Indigenous} {Languages} of the {Americas} ({AmericasNLP})}, pages 163--167, Toronto, Canada. Association for Computational Linguistics.

\bibitem[{Tang(2022)}]{tang_petci_2022}
Kenan Tang. 2022.
\newblock \href {http://arxiv.org/abs/2202.09509} {{PETCI}: {A} {Parallel} {English} {Translation} {Dataset} of {Chinese} {Idioms}}.
\newblock ArXiv:2202.09509 [cs].

\bibitem[{Tiedemann(2012)}]{tiedemann_parallel_2012}
Jörg Tiedemann. 2012.
\newblock Parallel data, tools and interfaces in {OPUS}.
\newblock In \emph{Proceedings of the {Eight} {International} {Conference} on {Language} {Resources} and {Evaluation}}, pages 2214--2218, Istanbul, Turkey. European Language Resources Association (ELRA).

\bibitem[{Touvron et~al.(2023)Touvron, Lavril, Izacard, Martinet, Lachaux, Lacroix, Rozière, Goyal, Hambro, Azhar, Rodriguez, Joulin, Grave, and Lample}]{touvron_llama_2023}
Hugo Touvron, Thibaut Lavril, Gautier Izacard, Xavier Martinet, Marie-Anne Lachaux, Timothée Lacroix, Baptiste Rozière, Naman Goyal, Eric Hambro, Faisal Azhar, Aurelien Rodriguez, Armand Joulin, Edouard Grave, and Guillaume Lample. 2023.
\newblock \href {http://arxiv.org/abs/2302.13971} {{LLaMA}: {Open} and {Efficient} {Foundation} {Language} {Models}}.
\newblock ArXiv:2302.13971 [cs].

\bibitem[{Vilar et~al.(2023)Vilar, Freitag, Cherry, Luo, Ratnakar, and Foster}]{vilar_prompting_2023}
David Vilar, Markus Freitag, Colin Cherry, Jiaming Luo, Viresh Ratnakar, and George Foster. 2023.
\newblock \href {https://aclanthology.org/2023.acl-long.859} {Prompting {PaLM} for {Translation}: {Assessing} {Strategies} and {Performance}}.
\newblock In \emph{Proceedings of the 61st {Annual} {Meeting} of the {Association} for {Computational} {Linguistics} ({Volume} 1: {Long} {Papers})}, pages 15406--15427, Toronto, Canada. Association for Computational Linguistics.

\bibitem[{Wang et~al.(2023)Wang, Lyu, Ji, Zhang, Yu, Shi, and Tu}]{wang_document-level_2023}
Longyue Wang, Chenyang Lyu, Tianbo Ji, Zhirui Zhang, Dian Yu, Shuming Shi, and Zhaopeng Tu. 2023.
\newblock \href {http://arxiv.org/abs/2304.02210} {Document-{Level} {Machine} {Translation} with {Large} {Language} {Models}}.
\newblock ArXiv:2304.02210 [cs].

\bibitem[{Wei et~al.(2023)Wei, Wei, Lin, Li, Zhang, Ren, Li, Wan, Cao, Xie, Hu, Li, Hui, Yu, Liu, Yang, Huang, and Xie}]{wei_polylm_2023}
Xiangpeng Wei, Haoran Wei, Huan Lin, Tianhao Li, Pei Zhang, Xingzhang Ren, Mei Li, Yu~Wan, Zhiwei Cao, Binbin Xie, Tianxiang Hu, Shangjie Li, Binyuan Hui, Bowen Yu, Dayiheng Liu, Baosong Yang, Fei Huang, and Jun Xie. 2023.
\newblock \href {http://arxiv.org/abs/2307.06018} {{PolyLM}: {An} {Open} {Source} {Polyglot} {Large} {Language} {Model}}.
\newblock ArXiv:2307.06018 [cs].

\bibitem[{Wicks and Post(2023)}]{wicks_identifying_2023}
Rachel Wicks and Matt Post. 2023.
\newblock \href {http://arxiv.org/abs/2311.02321} {Identifying {Context}-{Dependent} {Translations} for {Evaluation} {Set} {Production}}.
\newblock ArXiv:2311.02321 [cs].

\bibitem[{Wolf et~al.(2020)Wolf, Debut, Sanh, Chaumond, Delangue, Moi, Cistac, Rault, Louf, Funtowicz, Davison, Shleifer, Von~Platen, Ma, Jernite, Plu, Xu, Le~Scao, Gugger, Drame, Lhoest, and Rush}]{wolf_transformers_2020}
Thomas Wolf, Lysandre Debut, Victor Sanh, Julien Chaumond, Clement Delangue, Anthony Moi, Pierric Cistac, Tim Rault, Remi Louf, Morgan Funtowicz, Joe Davison, Sam Shleifer, Patrick Von~Platen, Clara Ma, Yacine Jernite, Julien Plu, Canwen Xu, Teven Le~Scao, Sylvain Gugger, Mariama Drame, Quentin Lhoest, and Alexander Rush. 2020.
\newblock \href {https://doi.org/10.18653/v1/2020.emnlp-demos.6} {Transformers: {State}-of-the-{Art} {Natural} {Language} {Processing}}.
\newblock In \emph{Proceedings of the 2020 {Conference} on {Empirical} {Methods} in {Natural} {Language} {Processing}: {System} {Demonstrations}}, pages 38--45, Online. Association for Computational Linguistics.

\bibitem[{Wu et~al.(2024)Wu, Tan, Meng, Stap, and Monz}]{wu2024far}
Di~Wu, Shaomu Tan, Yan Meng, David Stap, and Christof Monz. 2024.
\newblock How far can 100 samples go? unlocking overall zero-shot multilingual translation via tiny multi-parallel data.
\newblock \emph{arXiv preprint arXiv:2401.12413}.

\bibitem[{Xu et~al.(2023)Xu, Kim, Sharaf, and Awadalla}]{xu_paradigm_2023}
Haoran Xu, Young~Jin Kim, Amr Sharaf, and Hany~Hassan Awadalla. 2023.
\newblock \href {http://arxiv.org/abs/2309.11674} {A {Paradigm} {Shift} in {Machine} {Translation}: {Boosting} {Translation} {Performance} of {Large} {Language} {Models}}.
\newblock ArXiv:2309.11674 [cs].

\bibitem[{Yang et~al.(2023)Yang, Li, Zhang, and Zong}]{yang_bigtranslate_2023}
Wen Yang, Chong Li, Jiajun Zhang, and Chengqing Zong. 2023.
\newblock \href {http://arxiv.org/abs/2305.18098} {{BigTranslate}: {Augmenting} {Large} {Language} {Models} with {Multilingual} {Translation} {Capability} over 100 {Languages}}.
\newblock ArXiv:2305.18098 [cs].

\bibitem[{Zaninello and Birch(2020)}]{zaninello-birch-2020-multiword}
Andrea Zaninello and Alexandra Birch. 2020.
\newblock \href {https://aclanthology.org/2020.lrec-1.471} {Multiword expression aware neural machine translation}.
\newblock In \emph{Proceedings of the Twelfth Language Resources and Evaluation Conference}, pages 3816--3825, Marseille, France. European Language Resources Association.

\bibitem[{Zeng et~al.(2023)Zeng, Meng, Yin, and Zhou}]{zeng_tim_2023}
Jiali Zeng, Fandong Meng, Yongjing Yin, and Jie Zhou. 2023.
\newblock \href {http://arxiv.org/abs/2307.04408} {{TIM}: {Teaching} {Large} {Language} {Models} to {Translate} with {Comparison}}.
\newblock ArXiv:2307.04408 [cs].

\bibitem[{Zhang et~al.(2023)Zhang, Rajabi, Duh, and Koehn}]{zhang_machine_2023}
Xuan Zhang, Navid Rajabi, Kevin Duh, and Philipp Koehn. 2023.
\newblock Machine {Translation} with {Large} {Language} {Models}: {Prompting}, {Few}-shot {Learning}, and {Fine}-tuning with {QLoRA}.
\newblock In \emph{{WMT}}.

\bibitem[{Zhu et~al.(2024)Zhu, Chen, Zhang, Haddow, Shen, and Klakow}]{zhu2024fine}
Dawei Zhu, Pinzhen Chen, Miaoran Zhang, Barry Haddow, Xiaoyu Shen, and Dietrich Klakow. 2024.
\newblock Fine-tuning large language models to translate: Will a touch of noisy data in misaligned languages suffice?
\newblock \emph{arXiv preprint arXiv:2404.14122}.

\bibitem[{Zhu et~al.(2023{\natexlab{a}})Zhu, Liu, Dong, Xu, Huang, Kong, Chen, and Li}]{zhu_multilingual_2023}
Wenhao Zhu, Hongyi Liu, Qingxiu Dong, Jingjing Xu, Shujian Huang, Lingpeng Kong, Jiajun Chen, and Lei Li. 2023{\natexlab{a}}.
\newblock \href {http://arxiv.org/abs/2304.04675} {Multilingual {Machine} {Translation} with {Large} {Language} {Models}: {Empirical} {Results} and {Analysis}}.
\newblock ArXiv:2304.04675 [cs].

\bibitem[{Zhu et~al.(2023{\natexlab{b}})Zhu, Lv, Dong, Yuan, Xu, Huang, Kong, Chen, and Li}]{zhu_extrapolating_2023}
Wenhao Zhu, Yunzhe Lv, Qingxiu Dong, Fei Yuan, Jingjing Xu, Shujian Huang, Lingpeng Kong, Jiajun Chen, and Lei Li. 2023{\natexlab{b}}.
\newblock \href {http://arxiv.org/abs/2308.04948} {Extrapolating {Large} {Language} {Models} to {Non}-{English} by {Aligning} {Languages}}.
\newblock ArXiv:2308.04948 [cs].

\end{thebibliography}
\bibliographystyle{acl_natbib}

\appendix

\section{Details on experimental setup}
\label{sec:appendix_experimental_setup}
We run our fine-tuning experiments with the Hugging Face transformers library \citep{wolf_transformers_2020} and make use of DeepSpeed \citep{rasley_deepspeed_2020}.
We store intermediate checkpoints during fine-tuning to track how abilities evaluate over time.
We perform full fine-tuning on models up to 40B.
We use the AdamW optimizer with a cosine learning rate scheduler and 3\% warm-up percentage.
We empirically set the batch size to 128, learning rate to $2e-5$, and train for one epoch.
During inference we use beam search with a beam size of 4.
For LLaMA-65B, we fine-tune with QLoRA \citep{dettmers_qlora_2023}, using 8-bit quantization.
Following \citet{zhang_machine_2023} we set the low-rank approximation to 64 and the scaling factor for low-rank adaptation to 32.

\paragraph{Inference}
Depending on the evaluation set, we use a 0-shot or few-shot approach.
The prompt used for fine-tuning and 0-shot is shown in Table~\ref{tab:0_shot_prompt}.
For our few-shot setup, we find the 5 most similar source sentences and their corresponding target sentences from a corresponding train set (if available) or validation set.
The resulting prompt is displayed in Table~\ref{tab:5_shot_prompt}.
We use Sentence-BERT \citep{reimers_sentence-bert_2019} for encoding\footnote{We use \texttt{all-MiniLM-L6-v2} for English sentences and \texttt{paraphrase-multilingual-MiniLM-L12-v2} for non-English} and FAISS \citep{johnson_billion-scale_2019} for searching similar sentences.

\begin{table*}[t]
    \begin{center}
        \small
        \begin{tabular}{l}
        \toprule
        \texttt{Translate this from \{source\_language\} to \{target\_language\}:} \\
        \texttt{\{source\_language\}: \{source\_sentence\}} \\
        \texttt{\{target\_language\}: \{target\_sentence\}} \\
        \bottomrule
        \end{tabular}
    \end{center}
    \caption{Prompting template for fine-tuning and 0-shot inference. For fine-tuning \texttt{\{target\_sentence\}} is filled with the corresponding target sentence, and for 0-shot inference it is the empty string.}
    \label{tab:0_shot_prompt}
\end{table*}

\begin{table*}[t]
    \begin{center}
        \small
        \begin{tabular}{l}
        \toprule
        \texttt{Translate this from \{source\_language\} to \{target\_language\}:} \\
        \texttt{\{source\_language\}: \{source\_sentence$_1$\}} \\
        \texttt{\{target\_language\}: \{target\_sentence$_1$\}} \\
        \texttt{...} \\
        \texttt{\{source\_language\}: \{source\_sentence$_n$\}} \\
        \texttt{\{target\_language\}: } \\
        \bottomrule
        \end{tabular}
    \end{center}
    \caption{Prompting template for few-shot inference.}
    \label{tab:5_shot_prompt}
\end{table*}

\section{Additional results}
\label{sec:appendix_additional_results}

\subsection{General translation quality (WMT22)}
\begin{figure*}[ht]
    \centering
    \includegraphics[width=\linewidth]{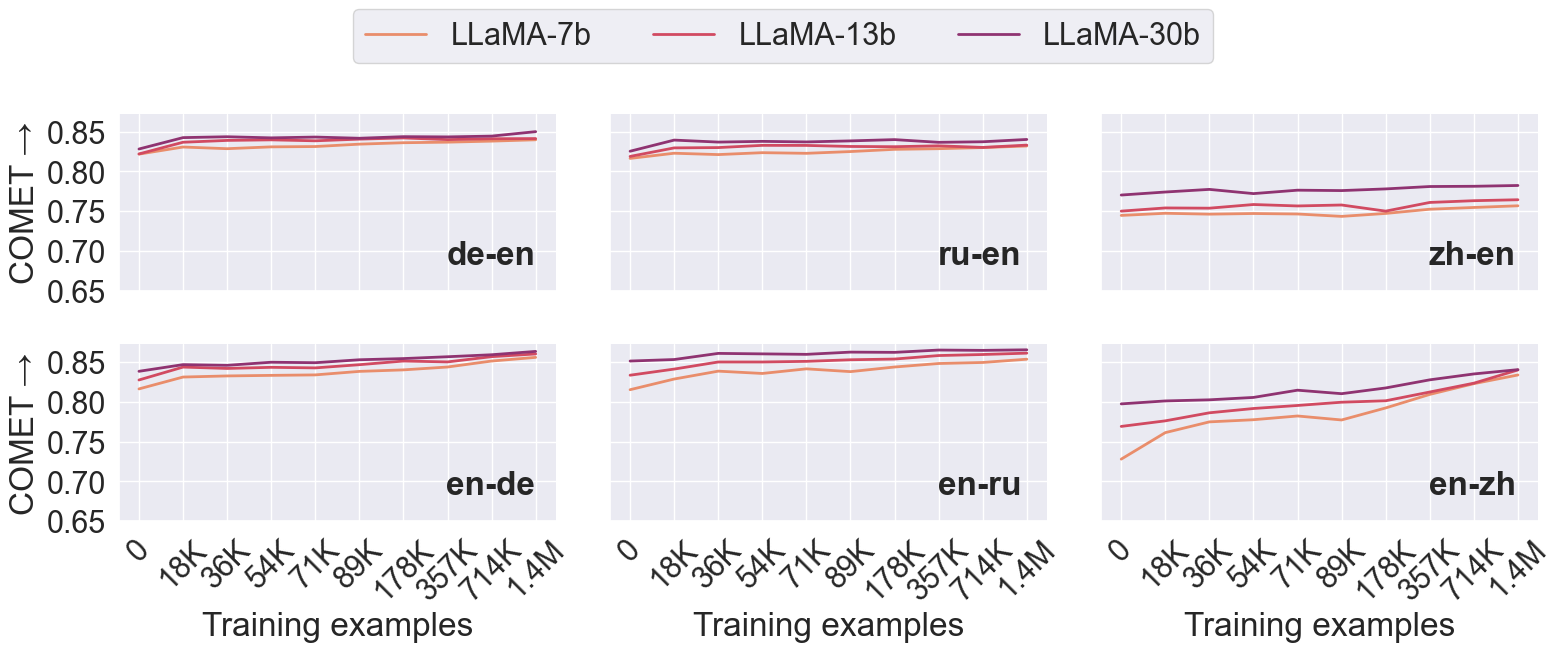}
    \caption{X$\rightarrow$English (top) and English$\rightarrow$X (bottom) COMET scores on WMT22 for different models trained on OPUS parallel data with different amounts of training data.
    }
    \label{fig:opus_general_quality_wmt}
\end{figure*}

Results on WMT22 for models trained on filtered web-crawled data are summarized in Figure~\ref{fig:opus_general_quality_wmt}.
Note that the data size up until 89K has relatively small increments, and later data sizes are doubled compared to the data point before it.
Generally, in contrast to training on human-written data, translation quality continues to increase when adding more data.

We compare the best checkpoints of models trained on human-written and OPUS data in Table \ref{tab:wmt_vs_opus_general_quality}.
In 15 out of 18 cases, models fine-tuned on the larger OPUS dataset results in better scores on WMT22.

\begin{table*}[th!]
    \centering
    \small
    \begin{tabular}{lllllll}
        \toprule
        & \multicolumn{2}{c}{de-en}
        & \multicolumn{2}{c}{ru-en}
        & \multicolumn{2}{c}{zh-en} \\
        \cmidrule(lr){2-3} 
        \cmidrule(lr){4-5}
        \cmidrule(lr){6-7}
        & \multicolumn{1}{l}{WMT} 
        & \multicolumn{1}{l}{OPUS}
        & \multicolumn{1}{l}{WMT} 
        & \multicolumn{1}{l}{OPUS}
        & \multicolumn{1}{l}{WMT}
        & \multicolumn{1}{l}{OPUS} \\
        
        \midrule
        LLaMA-7b         
            & $0.834$ (36K)
            & $\underline{0.840}$ (1.4M)
            & $0.831$ (18K)
            & $\underline{0.832}$ (1.4M)
            & $0.749$ (36K)
            & $\underline{0.757}$ (1.4M) \\                 

        LLaMA-13b
            & $0.842$ (18K)
            & $\underline{0.842}$ (178K)
            & $\underline{0.837}$ (18K)
            & $0.833$ (1.4M)
            & $\underline{0.765}$ (18K)
            & $0.764$ (1.4M) \\

        LLaMA-30b
            & $0.844$ (71K)
            & $\underline{0.850}$ (1.4M)
            & $\underline{0.843}$ (18K)
            & $0.840$ (1.4M)
            & $0.772$ (18K)
            & $\underline{0.782}$ (1.4M) \\
        \midrule \\
        & \multicolumn{2}{c}{en-de}
        & \multicolumn{2}{c}{en-ru}
        & \multicolumn{2}{c}{en-zh} \\
        \cmidrule(lr){2-3} 
        \cmidrule(lr){4-5}
        \cmidrule(lr){6-7}
        & \multicolumn{1}{l}{WMT} 
        & \multicolumn{1}{l}{OPUS}
        & \multicolumn{1}{l}{WMT} 
        & \multicolumn{1}{l}{OPUS}
        & \multicolumn{1}{l}{WMT}
        & \multicolumn{1}{l}{OPUS} \\
        \midrule
        LLaMA-7b         
            & $0.831$ (89K)
            & $\underline{0.856}$ (1.4M)
            & $0.840$ (71K)
            & $\underline{0.853}$ (1.4M)
            & $0.797$ (89K)
            & $\underline{0.834}$ (1.4M) \\                 

        LLaMA-13b
            & $0.843$ (54K)
            & $\underline{0.860}$ (1.4M)
            & $0.854$ (89K)
            & $\underline{0.861}$ (1.4M)
            & $0.818$ (89K)
            & $\underline{0.840}$ (1.4M) \\

        LLaMA-30b
            & $0.848$ (36K)
            & $\underline{0.863}$ (1.4M)
            & $0.862$ (18K)
            & $\underline{0.865}$ (1.4M)
            & $0.826$ (54K)
            & $\underline{0.840}$ (1.4M) \\

        \bottomrule
        \end{tabular}
        \caption{WMT22 COMET scores comparing models fine-tuned on human-written data (WMT) and filtered web-crawled data (OPUS).
        Parentheses indicate the number of fine-tuning examples seen by the best-performing checkpoints.
        Best scores are \underline{underlined}.}
        \label{tab:wmt_vs_opus_general_quality}
\end{table*}

\subsection{Formality steering}
\begin{figure}[ht]
    \centering
    \includegraphics[width=\linewidth]{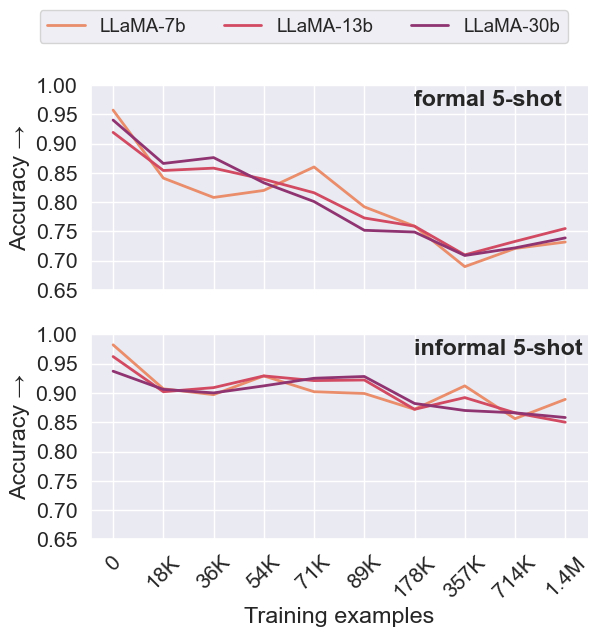}
    \caption{Accuracy of formality markers for models trained on OPUS data.
    }
    \label{fig:opus_formality_steering}
\end{figure}

Figure \ref{fig:opus_formality_steering} shows that the ability to generate correct formality markers also degrades when fine-tuning on OPUS data.
Extended fine-tuning continues to show benefits in terms of general translation quality (see Figure \ref{fig:opus_general_quality_wmt}), but the ability to generate correct formality markers continues to degrade.

\subsection{Technical domains}
\begin{figure*}[ht]
    \centering
    \includegraphics[width=\linewidth]{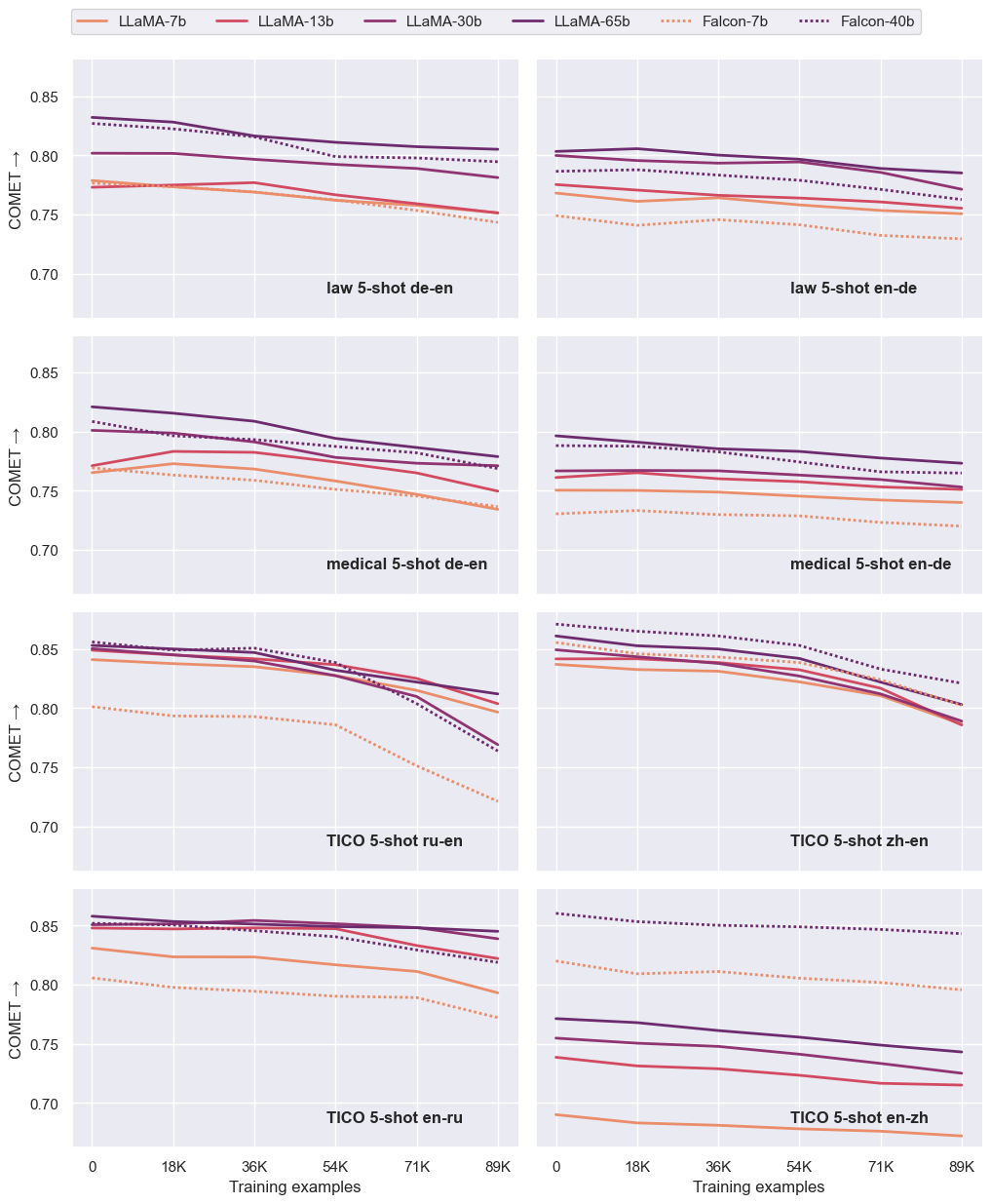}
    \caption{COMET on technical domains using 5-shot examples for models trained on human-written translations. 
    }
    \label{fig:wmt_technical_domains_full}
\end{figure*}

Figure \ref{fig:wmt_technical_domains_full} shows the outcome of fine-tuning on human-written data across all evaluated domains and language directions.
We observe a consistent trend: fine-tuning impairs the few-shot technical translation capabilities, and generally further fine-tuning results in more degradations.
The COMET scores correlate negatively with datastore size ($\rho=-0.27$, $p<0.001$), indicating that fine-tuning on more data results in larger degradations.

\begin{figure*}[ht]
    \centering
    \includegraphics[width=\linewidth]{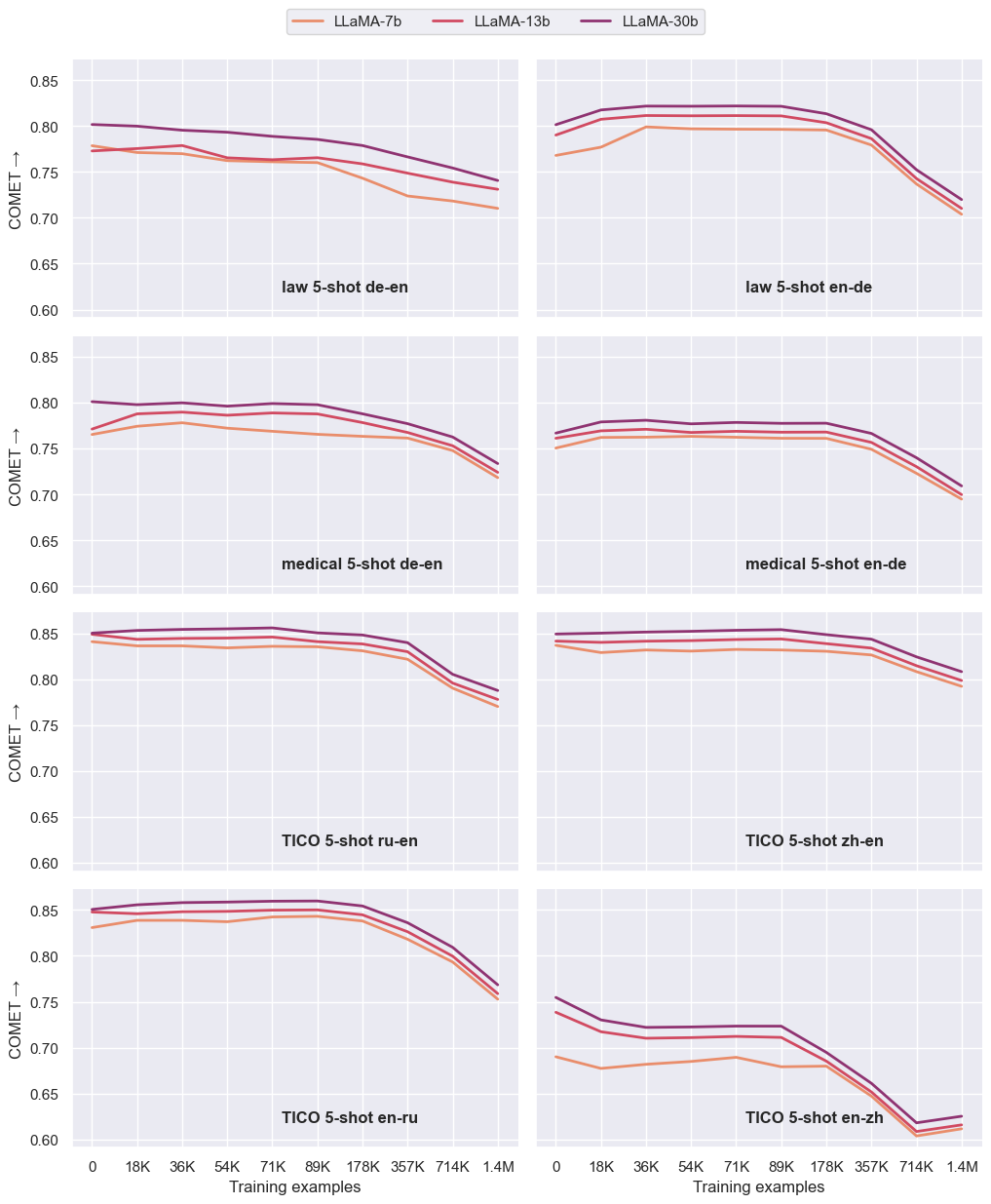}
    \caption{COMET on technical domains using 5-shot examples for models trained on OPUS data. 
    }
    \label{fig:opus_technical_domains_full}
\end{figure*}

The effects of fine-tuning are also analyzed using filtered web-scraped data from OPUS, as shown in Figure \ref{fig:opus_technical_domains_full}.
Similar to the previous findings, an increase in data volume for fine-tuning corresponds to performance degradations, evidenced by a negative correlation between COMET scores and datastore size ($\rho=-0.33$, $p<0.001$).

However, OPUS data reveals that these degradations manifest more gradually compared to the WMT dataset.
This discrepancy is likely due to OPUS's broader domain coverage, in contrast to the specialized news content of the human-curated WMT dataset.
Notably, while fine-tuning on OPUS data leads to deterioration in technical domain translation accuracy when leveraging few-shot examples, it concurrently continues to enhance overall translation quality (Figure \ref{fig:opus_general_quality_wmt}), underscoring a nuanced impact of fine-tuning across different data types and translation tasks.

e
\subsection{Contextualization of document-level input}

\begin{figure}[ht]
    \centering
    \includegraphics[width=\linewidth]{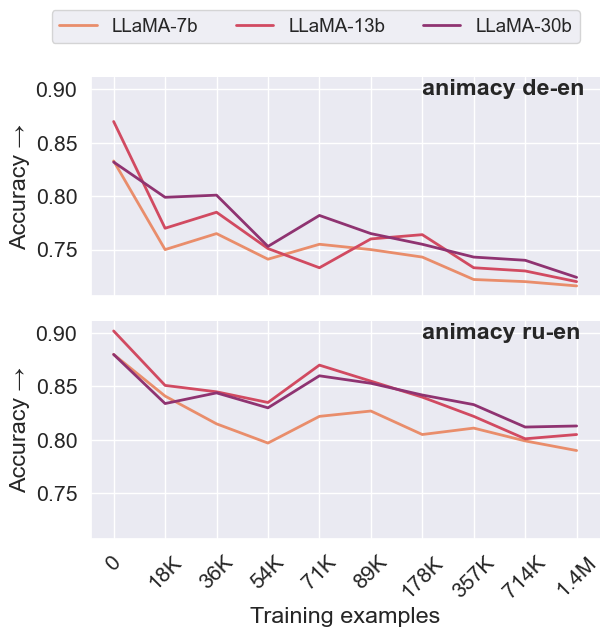}
    \caption{Accuracy of animacy contextualization for German$\rightarrow$English and Russian$\rightarrow$English for models fine-tuned with human-written translations.
    }
    \label{fig:opus_animacy}
\end{figure}

Figure \ref{fig:opus_animacy} shows that the animacy contextualization accuracy of document-level input degrades for models fine-tuned on filtered web-crawled OPUS data.
We observe a negative correlation between accuracy and fine-tuning dataset size ($\rho=-0.49$, $p<0.001$).

\end{document}